\documentclass[conference]{IEEEtran}
\IEEEoverridecommandlockouts
\let\labelindent\relax
\usepackage{enumitem}
\usepackage{ulem}
\usepackage{nicefrac}
\usepackage{siunitx}
\usepackage{array,framed}
\usepackage{amsthm}
\usepackage{booktabs}
\usepackage{
  color,
  float,
  epsfig,
  wrapfig,
  graphics,
  graphicx,
  subcaption
}
\usepackage{etoolbox}
\usepackage{textcomp,amssymb}
\usepackage{setspace}
\usepackage{latexsym,fancyhdr,url}
\usepackage{enumerate}
\usepackage{algorithm}
\usepackage[noend]{algpseudocode}
\usepackage{graphics}
\usepackage{xparse} 
\usepackage{xspace}
\usepackage{multirow}
\usepackage{csvsimple}
\usepackage{balance}

\usepackage{
  tikz,
  pgfplots,
  pgfplotstable
}
\usepackage{hyperref}

\usetikzlibrary{
  shapes.geometric,
  arrows,
  external,
  pgfplots.groupplots,
  matrix
}
\pgfplotsset{compat=1.18} 

\theoremstyle{plain}

\theoremstyle{definition}
\newtheorem{definition}{Definition}
\usepackage{mathtools}
\usepackage{cite}
\usepackage{amsmath,amssymb,amsfonts}
\usepackage{graphicx}
\usepackage{textcomp}
\usepackage{xcolor}
\def\BibTeX{{\rm B\kern-.05em{\sc i\kern-.025em b}\kern-.08em
    T\kern-.1667em\lower.7ex\hbox{E}\kern-.125emX}}

\makeatletter
\newenvironment{breakablealgorithm}
  {
   \begin{center}
     \refstepcounter{algorithm}
     \hrule height.8pt depth0pt \kern2pt
     \renewcommand{\caption}[2][\relax]{
       {\raggedright\textbf{\ALG@name~\thealgorithm} ##2\par}%
       \ifx\relax##1\relax 
         \addcontentsline{loa}{algorithm}{\protect\numberline{\thealgorithm}##2}%
       \else 
         \addcontentsline{loa}{algorithm}{\protect\numberline{\thealgorithm}##1}%
       \fi
       \kern2pt\hrule\kern2pt
     }
  }{
     \kern2pt\hrule\relax
   \end{center}
  }
\makeatother
\begin{document}

\title{Krait: A Backdoor Attack Against Graph Prompt Tuning}
\author{\IEEEauthorblockN{Ying Song}
\IEEEauthorblockA{\textit{University of Pittsburgh}\\
yis121@pitt.edu}
\and
\IEEEauthorblockN{Rita Singh}
\IEEEauthorblockA{\textit{Carnegie Mellon University}\\
rsingh@cs.cmu.edu}
\and
\IEEEauthorblockN{Balaji Palanisamy}
\IEEEauthorblockA{\textit{University of Pittsburgh}\\
bpalan@pitt.edu}}


\maketitle
\begin{abstract}


Graph prompt tuning has emerged as a promising paradigm to effectively transfer general graph knowledge from pre-trained models to various downstream tasks, particularly in few-shot contexts. However, its susceptibility to backdoor attacks, where adversaries insert triggers to manipulate outcomes, raises a critical concern. We conduct the first study to investigate such vulnerability, revealing that backdoors can disguise benign graph prompts, thus evading detection. We introduce Krait, a novel graph prompt backdoor. Specifically, we propose a simple yet effective model-agnostic metric called label non-uniformity homophily to select poisoned candidates, significantly reducing computational complexity. To accommodate diverse attack scenarios and advanced attack types, we design three customizable trigger generation methods to craft prompts as triggers. We propose a novel centroid similarity-based loss function to optimize prompt tuning for attack effectiveness and stealthiness. Experiments on four real-world graphs demonstrate that Krait can efficiently embed triggers to merely 0.15\% to 2\% of training nodes, achieving high attack success rates without sacrificing clean accuracy. Notably, in one-to-one and all-to-one attacks, Krait can achieve 100\% attack success rates by poisoning as few as 2 and 22 nodes, respectively. Our experiments further show that Krait remains potent across different transfer cases, attack types, and graph neural network backbones. Additionally, Krait can be successfully extended to the black-box setting, posing more severe threats. Finally, we analyze why Krait can evade both classical and state-of-the-art defenses, and provide practical insights for detecting and mitigating this class of attacks.

\end{abstract}


\section{Introduction}
\label{sec:intro} 



Graph neural networks (GNNs) have achieved exceptional success in various real-world applications, such as recommendation systems \cite{recomm_app}, anomaly detection \cite{anomaly_detect}, drug discovery, and disease diagnosis \cite{bio_health}. Nevertheless, traditional GNNs heavily rely on manual annotations and are prone to over-fitting, especially when testing graphs are out-of-distribution. To address these issues, the ``Graph pre-training and fine-tuning" paradigm has been widely employed. It pre-trains GNNs on diverse graph datasets and then transfers the learned graph knowledge to downstream tasks with minimal adaptation steps. Despite its success, fine-tuning large-scale GNNs can be time-consuming and resource-intensive. Additionally, pre-training tasks can misalign with downstream tasks in terms of semantic knowledge, learning objectives, and task difficulty, leading to negative transfers \cite{all_in_one,graph_prompt_survey}. Recently, the ``Graph pre-training and prompt-tuning" \cite{GPPT,graph_prompts,all_in_one} has emerged as a promising solution to mitigate these misalignments. It utilizes graph prompts as free parameters to reformulate downstream tasks into pre-training tasks, thereby reducing computational overhead and adaptation complexity.

\noindent\textbf{Graph Backdoor.} Previous research has shown that GNNs are vulnerable to graph backdoors \cite{graph_backdoor, unnoticeable_graph_backdoor, multi_gnn_backdoor}, which implant triggers into graphs so that GNNs misclassify nodes attached with triggers as a target label while maintaining the accuracy of clean nodes. However, existing trigger generation methods are often random or computationally intensive \cite{random_backdoor, graph_backdoor, explain_backdoor}, reducing the efficacy of backdoor attacks. Additionally, Dai et al. \cite{unnoticeable_graph_backdoor} empirically demonstrate that most graph backdoors consume large attack budgets, i.e., poisoning rate (PR), to guarantee attack effectiveness on large-scale graphs, but their generated triggers can be easily detected by homophily and node-similarity-based defense mechanisms. Besides, existing studies primarily focus on one-to-one attacks, with more complex all-to-one and all-to-all attacks less explored. Although a few efforts have been made to design one-to-many and many-to-one attacks \cite{multi_gnn_backdoor, one_to_many_backdoor}, the word ``many" in these contexts only refers to a few labels. Most importantly, to the best of our knowledge, backdoor attacks against graph prompt tuning are yet to be explored. However, prior work cannot be trivially adapted into graph prompt tuning since the majority of them are designed for traditional GNNs \cite{unnoticeable_graph_backdoor, random_backdoor, explain_backdoor, multi_gnn_backdoor}. Even though several studies support backdooring pre-trained GNNs, poisoning the pre-training graphs can be time-consuming, and perturbing pre-trained GNNs \cite{graph_backdoor, pre-train_backdoor, contrastive_backdoor} will inherit the limitations of the pre-trained models themselves. Due to privacy protection and proprietary concerns, many companies and institutions cannot open-source their pre-training datasets along with the pre-training process, making it infeasible to poison the pre-training phase of graph prompt tuning.

\noindent\textbf{Graph Prompt Backdoor.} We observe an interesting phenomenon: graph prompt tuning and graph backdoor attacks both append small-scale subgraphs to the input graphs, but serve for inverse purposes. Graph prompt tuning treats subgraphs as graph prompts to improve the performance of downstream tasks, while graph backdoor attacks implant subgraphs as triggers to selected poisoned candidates, ensuring they are classified as the desired label \cite{graph_backdoor,unnoticeable_graph_backdoor}. This inherent similarity motivates us to investigate the following three questions: 

 1) \uline{Is graph prompt tuning susceptible to backdoor attacks?} 
 
 2) \uline{If so, can attackers craft graph prompts as triggers and exploit this vulnerability to facilitate graph backdoor attacks?}
 
 3) \uline{How can we design an effective and stealthy graph prompt backdoor for various attack types?} 

Notably, designing a graph prompt backdoor poses additional challenges. First, graph prompt tuning adapts pre-trained GNNs to downstream tasks within a few epochs, which might not establish a solid mapping between triggers and target labels, thus compromising the effectiveness of backdoor attacks. Second, disguising triggers as benign graph prompts requires attackers to generate adaptive triggers, otherwise, these triggers can be destroyed during the prompt tuning process.

\noindent\textbf{Practicability and Social Impact.} Recently, an open-source library called ProG\footnote{\url{https://github.com/sheldonresearch/ProG/tree/main}} has been released to the public, which packs many state-of-the-art (SOTA) graph prompt tuning models \cite{graph_prompt_survey}. Due to its flexibility, it is anticipated that more developers can tailor this library to the specific requirements of their APIs or platforms. However, developers can become malicious model providers, who may poison the customized downstream graph with well-crafted triggers before model training, or modify handcrafted or auto-generated graph prompts to implement graph backdoors. Additionally, malicious users can manipulate the handcrafted graph prompts and inject them through user interfaces, or even send malicious instructions to generate graph prompts for poisoned samples. In high-stake scenarios such as healthcare \cite{healthcare_graph_prompt}, attackers could mislead such APIs or platforms to classify patients who should be diagnosed with illnesses as non-disease cases, potentially delaying critical treatments and endangering lives. Similarly, in less sensitive areas such as recommendation systems \cite{prompt_recommend}, attackers could manipulate the systems to expose more inappropriate or harmful content, leading to severe social impacts such as exacerbating discrimination, fostering hatred, and aiding criminal activities. We bring these critical security concerns into the research community, hoping that our work will stimulate future research on detecting and mitigating graph prompt backdoors. 


\noindent\textbf{Our Proposal.} In this paper, we conduct the first study to answer the first and second questions.
Our observations reveal consistent patterns between graph prompt tuning and graph backdoors, highlighting the inherent vulnerability to crafting graph prompts as triggers. To answer the third question, we introduce Krait (Bac\underline{k}doo\underline{r} ag\underline{ai}nst graph prompt \underline{t}uning), a novel graph prompt backdoor. We design a model-agnostic metric, label non-uniformity homophily, to identify the most vulnerable poisoned candidates before model training, largely reducing computational costs. 
Next, we develop three trigger generation methods for various attack scenarios and advanced attack types under the more stringent white-box setting. These triggers are then adaptively tuned along with benign prompts, thus avoiding detection and destruction. 
Additionally, we incorporate a novel centroid similarity-based loss to improve the effectiveness and stealthiness of Krait. Finally, we extend Krait to the black-box setting, enhancing its practicality.

\noindent\textbf{Contributions.} Our five key contributions are as follows:
\begin{itemize}[leftmargin=*]

\item We propose two node-level homophily metrics and three centroid similarity-based metrics to analyze the behaviors of graph prompt tuning before and after backdoor injections. The observed patterns unveil the intrinsic vulnerability to treating graph prompts as triggers.
\item We introduce Krait, a novel graph prompt backdoor attack with three key components: 1) a model-agnostic metric to identify the most vulnerable poisoned candidates; 2) three versatile trigger generation methods to generate graph prompts as triggers; 3) a novel centroid similarity-based loss function to boost the attack's effectiveness and stealthiness.
\item Our extensive experiments show that Krait can generate effective and covert triggers across various transfer cases, attack types, trigger generation methods, and GNN backbones under the more stringent white-box setting. 
We strictly set PR as 0.15\%-2\% for different attack types. In some specific cases, Krait only implants minimal-scale triggers into two nodes but achieves a 100\% attack success rate (ASR).
\item We successfully extend Krait to the black-box setting.
\item We demonstrate that existing defenses cannot effectively detect and mitigate Krait, highlighting the urgent need for more reliable defenders against graph prompt backdoors. We further provide general suggestions for researchers and practitioners to safeguard graph prompt tuning models.
\end{itemize}

\vspace{-2mm}
\section{Background}
\label{sec:background}

\subsection{Notations}
Given an undirected attributed graph $\mathcal{G}=(\mathcal{V}, \mathcal{E}, X)$, $\mathcal{V}$ denotes a node set with $|\mathcal{V}|$ nodes and each node $v$ has a feature vector $X_v\in\mathcal{R}^{1\times d}$, where $d$ is the dimension of node features, $\mathcal{E}$ represents an edge set with $|\mathcal{E}|$ edges, which can be captured by the adjacency matrix $A\in\mathcal{R}^{|\mathcal{V}|\times|\mathcal{V}|}$ and $A_{u,v}=1$ i.i.f. the edge $(u,v)\in\mathcal{E}$. Hereby, we reform $\mathcal{G}=(A, X)$.

\subsection{Graph Neural Networks}
In essence, the remarkable performance of the majority of GNNs root in message-passing mechanisms, which aggregate each node $v\in\mathcal{V}$'s information from its local neighborhood $\mathcal{N}(v)$ and further update its node embedding $H^{l}_{v}$ after $l$ iterations. Formally, this process can be expressed as:
\begin{equation}
    H_{v}^{l}=UPD^{l}(H_{v}^{l-1},AGG^{l-1}(\{H_{u}^{l-1}:u\in\mathcal{N}(v)\}))
\end{equation}
where $H_{v}^{0}=X_v$, and $UPD$ and $AGG$ are two arbitrary differentiable functions to design diverse GNN variants. For node classification tasks, generally, $H_{v}^{l}$ is fed into a linear classifier with a softmax function for final prediction. As for graph classification tasks, readout functions can be deployed to encode graph-level representation by $Z_{\mathcal{G}}=REA({H_{u}^{l}|u\in\mathcal{V}})$, and $REA$ can be a simple permutation invariant function, e.g., average, summation or a more sophisticated graph pooling operation. Since transferring graph-level knowledge is more effective and generalizable, we reformulate node-level tasks into graph-level tasks by constructing $k$-hop ego-networks $\mathcal{G}_{s(v)}$ for $v\in \mathcal{V}$ \cite{all_in_one}, where the $k$-hop ego-network denotes the induced subgraph of the target node $v$ and the shortest path distance from any other node to $v$ is set to $k$.

\vspace{-2mm}
\subsection{Graph Pre-training and Prompt-tuning}

\noindent\textbf{Pre-training Phase.} Graph pre-training transfers general graph knowledge to downstream tasks without requiring manual annotations. As indicated in the previous work\cite{graph_prompt_survey}, existing graph pre-training methods can be categorized into two main branches: graph generative learning and graph contrastive learning (GCL)-based methods. We solely consider the latter throughout this paper. Generally, GCL-based methods exploit graph augmentation techniques to generate two augmented views $\hat{\mathcal{G}}_{s(v),1}$ and $\hat{\mathcal{G}}_{s(v),2}$ for each $\mathcal{G}_{s(v)}$ as positive pairs, where $\hat{\mathcal{G}}_{s(v)}\sim q(\hat{\mathcal{G}}_{s(v)}|\mathcal{G}_{s(v)})$, and $q(\cdot|\mathcal{G}_s(v))$ denotes the augmentation distribution conditioned on $\mathcal{G}_{s(v)}$, $v\in \mathcal{V}$; negative pairs $\hat{\mathcal{G}}_{{s(v)}^{\prime}}$ are generated from the rest $|\mathcal{V}|-1$ augmented graphs \cite{graphcl}. GCL-based graph pre-training aims to minimize the contrastive loss to pre-train the GNN model $f_{pre}$:
\begin{equation}
    \min\limits_{\theta_{pre}}\mathcal{L}_{pre}=\sum\limits_{v\in\mathcal{V}}log\frac{-exp(sim(Z_{\mathcal{G}_{s(v),1}}, Z_{\mathcal{G}_{s(v),2}})/\tau)}{\sum_{v\prime=1,v\prime \neq v}exp(sim(Z_{\mathcal{G}_{s(v),1}}, Z_{\mathcal{G}{s(v)\prime,2}})/\tau)}
\end{equation}
where $\tau$ is the temperature parameter. 

\begin{wrapfigure}{r}[0cm]{0pt}  
    \vspace{-6mm}
    \includegraphics[width=3.6cm]{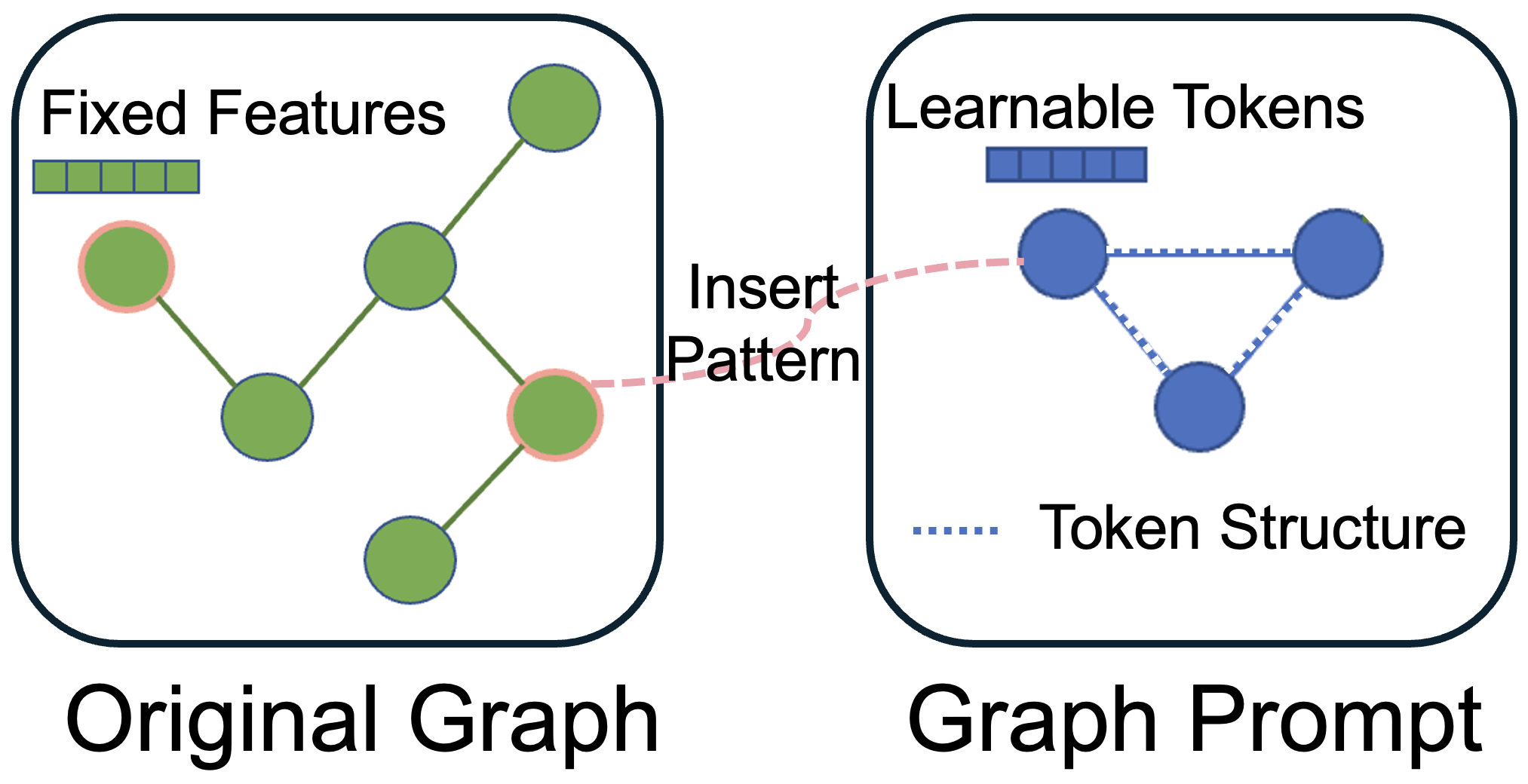} 
    \caption{Graph prompt}
    \vspace{-8mm}
\end{wrapfigure}

\noindent\textbf{Prompt-tuning Phase.}
Graph prompts generally contain three components: prompt tokens, token structures, and inserting patterns \cite{graph_prompt_survey}. Prompt tokens are usually learnable vectors. Token structures denote how each prompt token links with the rest and inserting patterns define how to attach prompts to original graphs. The simplest graph prompts are additional feature vectors concatenated to existing node features, while more advanced designs consider the inner structures of prompt tokens and their integration with the input graphs \cite{all_in_one}. Graph prompt tuning flexibly utilizes such graph prompts to reformulate downstream tasks into the pre-training task space \cite{graph_prompt_survey, all_in_one}. As for prompt generation, the graph prompt tuning model $f_{prm}$ employs a graph prompt generator $g_{\psi}:\mathcal{G}\rightarrow\mathcal{G}$ to transform each ego-network $\mathcal{G}_{s(v)}$ into the prompted subgraph $g_{\psi}(\mathcal{G}_{s(v)})=\mathcal{G}_{s(v)}\oplus\mathcal{G}_{\Delta}$, where $\mathcal{G}_{\Delta}$ are learnable graph prompts with prompt tokens $\mathcal{P}$ and token structures $\mathcal{E}_{\Delta}$, and $\oplus$ denotes inserting patterns. As for task-specific tuning, given a frozen pre-trained GNN $f_{pre}$, $f_{prm}$ strives to minimize the classification loss between final prediction $\hat{Y}$ and ground-truth label $Y$, which can be formulated as:
\begin{equation}
    \min\limits_{\psi,\theta_{prm}}\mathcal{L}_{prm}=\sum\limits_{v\in\mathcal{V}}l(f_{prm}(g_{\psi}(\mathcal{G}_{s(v)}), \theta_{nor}), Y_i)
\label{prm}
\end{equation}
where $\theta_{nor}$ contains parameters from the graph prompt tuning model $\theta_{prm}$ and frozen parameters from the pre-trained model $\theta_{pre}$. $l(\cdot)$ can be any loss function, e.g. entropy loss. 

\vspace{-3mm}
\subsection{Graph Backdoor}
Different from existing graph backdoors \cite{graph_backdoor} that inject triggers to the pre-trained GNN model $f_{pre}$, our attackers endeavor to generate a universal trigger $\delta$ and embed it to the graph prompt tuning model $f_{prm}$. They aim to misclassify the target nodes attached with triggers as the desired target label, and meanwhile, maintain a high classification accuracy of clean samples. which can be expressed as:
\begin{equation} 
\begin{split}
    \min\limits_{\theta_{tri}}\mathcal{L}_{tri}&= \sum\limits_{v\in\mathcal{V}_{c}}l(f_{prm}(g_{\psi}(\mathcal{G}_{s(v)}), \theta), Y_v)+ \\    &\phantom{=\;\;}\sum\limits_{u\in\mathcal{V}_{p}}l(f_{prm}(g_{\psi}(\mathcal{G}_{s(u)})\oplus\delta, \theta), Y_t)
\end{split}
\label{backdoor_express}
\end{equation}
where $Y_t$ is the desired target label and $\theta$ combines $\theta_{nor}$ and $\theta_{tri}$. $\mathcal{V}_{c}$ and $\mathcal{V}_{p}$ denote clean and poisoned node sets.

\vspace{-2mm}
\section{Empirical Analysis and Problem Statement}
\label{sec:prelim_exp}


In this section, we aim to answer the first and second research questions to show the inner vulnerability of graph prompt tuning and how it can be exploited by backdoor attacks. We design two types of metrics to investigate the behaviors of graph prompt tuning before and after backdoor injections. Our experimental results demonstrate that graph prompt tuning is vulnerable to backdoor attacks and exposes a new attack channel for disguising triggers as benign graph prompts. These findings motivate us to develop more effective and stealthy backdoor attacks against graph prompt tuning. Finally, we formalize our research problem in a new subsection.


We use All-in-One, the state-of-the-art (SOTA) graph prompt tuning model \cite{all_in_one}, to conduct our experiments. All-in-One supports the customization of downstream graphs, making it more flexible and practical for real-world applications. Additionally, it provides a universal framework for graph prompt tuning, thus enhancing its generalizability. Notably, many other graph prompt tuning models \cite{GPPT,VNT} can be viewed as special instances of All-in-One \cite{graph_prompt_survey}.


We consider two transfer settings, within-domain transfer and out-of-domain transfer. We employ four real-world datasets from diverse domains, namely, Cora, CiteSeer \cite{cora_citeseer}, Physics, and Computers \cite{physics_computers}. Please refer to Appendix \ref{appendix_dataset} for detailed graph description and statistics.

\vspace{-5mm}
\subsection{Metric Design}
We first introduce two node-centric homophily metrics and three centroid similarity metrics to investigate the behaviors of graph prompt tuning before and after backdoor injections.

\subsubsection{\textbf{Node-centric Homophily}}
Homophily describes the phenomenon where connected nodes tend to share similar labels or features \cite{unnoticeability_node_centric_homo}. Since many real graphs are homophilic and most existing GNNs implicitly follow homophily assumptions, we leverage homophily to investigate the behaviors of graph prompt tuning before and after backdoor injections.

However, graph prompt tuning generally constructs an ego-network for each targeted node, which involves transforming nodes into local subgraphs. Each ego-network is paired with a specific node label, but the nodes within the ego-network do not have labels, which is also the case with the graph prompt appended to each ego-network. Therefore, we cannot directly calculate label-based homophily. To address this issue, based on node-centric homophily \cite{unnoticeability_node_centric_homo}, we propose two metrics---local-subgraph homophily and global-view homophily---to compute node feature similarity between a node and its $1$-hop neighbors in different ways. 

\definition (Local-subgraph Homophily). We first compute the node-centric homophily for each node $u\in\mathcal{G}_{s(v)}$ and then take the average to represent the subgraph-level homophily of $\mathcal{G}_{s(v)}$, i.e., the local homophily of each node $v\in\mathcal{G}$, which can be formally expressed as: 
\begin{equation}
    h_v=\frac{1}{|\mathcal{V}_{s(v)}|}\sum\limits_{u\in\mathcal{G}_{s(v)}}sim(r_u, X_u), r_u=\sum\limits_{w\in\mathcal{N}(u)}\frac{1}{\sqrt{d_u d_w}}X_w
\end{equation}
where $|\mathcal{V}_{s(v)}|$ is the node size of the ego-network $\mathcal{G}_{s(v)}$, $d_u$ is the degree of node $u$, and $sim(\cdot)$ is the predefined similarity metric, e.g., cosine similarity. 

\definition (Global-view Homophily). Trivially averaging all node features in $\mathcal{G}_{s(v)}$ ignores its sub-topological structure $A_{s(v)}$. Therefore, we first acquire the subgraph embedding $Z_{\mathcal{G}_{s(v)}}$ as new features for each node $v\in\mathcal{G}$, and then compute the node-centric homophily based on the global topology $A$ of the original graph $\mathcal{G}$. The formal expression is as follows:
\begin{equation}
    h_v=sim(t_v, Z_{\mathcal{G}_{s(v)}}), t_v=\sum\limits_{u\in\mathcal{N}(v)}\frac{1}{\sqrt{d_v d_u}}Z_s(u)
\end{equation}

\subsubsection{\textbf{Centroid Similarity}} 
For $\mathcal{G}$ with a label set $|Y|$, $\mathcal{G}_{s(v)}$ is associated with $Y_v\in |Y|$, label centroids can be computed as: $L_{j}=\frac{1}{\sum\mathbb{I}(Y_v=j)}\sum Z_{\mathcal{G}_{s(v)}}$, where $j\in |Y|$, and $\mathbb{I}(\cdot)$ is an indicator function. 

\definition (Centroid Alignment, Misalignment, and Difference). Intuitively, given any label pair$(Y_v,j), Y_v\neq j$, for $\mathcal{G}_{s(v)}$, the alignment between $Z_{\mathcal{G}_{s(v)}}$ and the corresponding label centroid $L_{Y_v}$ should be larger, and the alignment between $Z_{\mathcal{G}_{s(v)}}$ and $L_{j}$ should be smaller. Therefore, we define the centroid alignment and misalignment as:
\begin{equation}
    \mathcal{CA}_v=sim(Z_v,L_{Y_v}); \mathcal{CM}_i=sim(Z_v,L_{j})
\end{equation}
Accordingly, the centroid difference is calculated as the difference between the centroid alignment and the misalignment, which can be expressed as  $\mathcal{CF}_{v}=\mathcal{CA}_{v}-\mathcal{CM}_{v}$.

Notably, $\mathcal{G}_{s(v)}$ can be easily replaced with the prompted subgraph $g_{\psi}(\mathcal{G}_{s(v)})$ once graph prompt tuning is applied.

\vspace{-2mm}
\subsection{Attack Setting}

We follow the same setting in Yang et.al. \cite{pre-train_backdoor} to choose a target label with the smallest number of training nodes and select the victim label with the largest number of training nodes in the downstream graph. 
We will consider more practical strategies to select target labels for different attack types in Section \ref{sec:framework}. Due to space consideration, the attack pipeline is included in Appendix \ref{attack_pipe}.

Notably, our attack setting differs from the more stringent and practical one in Section \ref{sec:framework}. In this section, we aim to demonstrate that graph prompt tuning is vulnerable to backdoor attacks and investigate how it behaves when ASRs are nearly $100\%$, regardless of attack stealthiness. 

\vspace{-2mm}
\subsection{Visualization and Analysis}



Since we focus solely on one-to-one attacks, we visualize and analyze the node-centric homophily and centroid similarity results of poisoned nodes with the same target label. To better understand the changes occurring before and after graph prompt tuning, and with and without trigger injection, we also visualize embedding spaces across these four scenarios.

\subsubsection{\textbf{Node-centric Homophily}}
Due to space constraints, we only exhibit two types of node-centric homophily distributions for poisoned candidates in the CiteSeer-to-Cora transfer case. 


\begin{figure}[htbp]
    \centering
    \vspace{-2mm}
    \includegraphics[width=\linewidth]{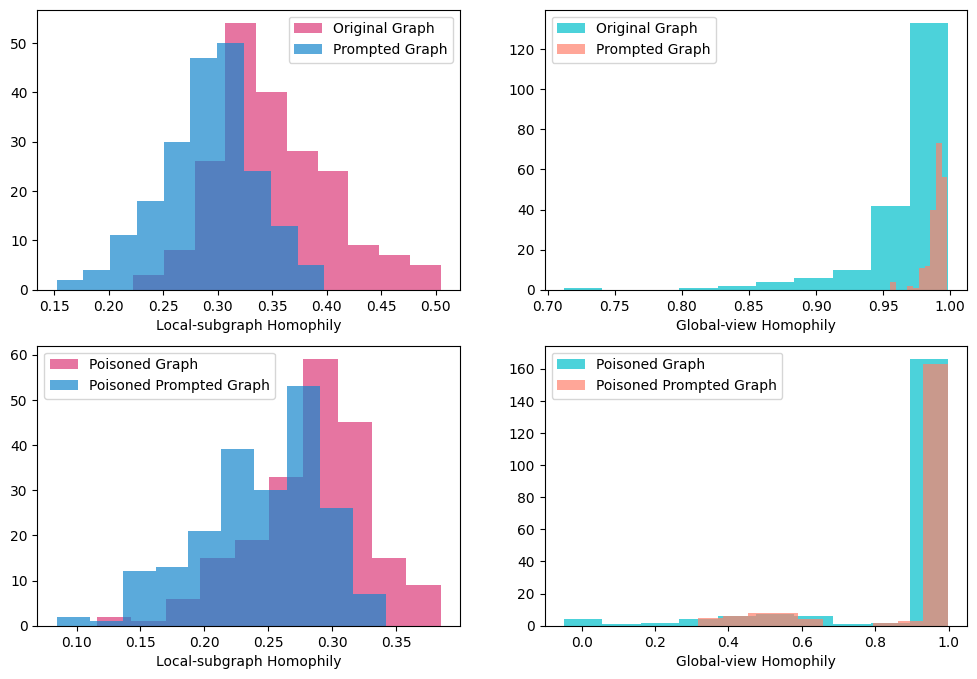}
    \caption{Local-subgraph (LHS) and global-view (RHS) homophily distributions of poisoned nodes. In the top two subfigures, pink and green denote original graphs, while blue and orange represent prompted graphs. In the bottom subfigures, pink and green show poisoned graphs, and blue and orange depict poisoned prompted graphs.}
    \label{homo_bkd}
    \vspace{-8mm}
\end{figure}


\noindent\textbf{Takeaways.}
From Figure \ref{homo_bkd}, we observe that after benign graph prompt tuning, the local-subgraph homophily of poisoned nodes generally decreases, while the global-view homophily tends to increase. We observe the same trends in graph backdoor attacks. Notably, whether the graph prompt tuning is benign or backdoored, it does not significantly alter homophily distributions. These findings hold when generalized to the whole raining graph and other positive transfers. Conversely, in the negative transfer, after graph prompt tuning, both the global-view homophily of the overall graph and that of the poisoned nodes significantly decrease. For further details, please refer to Appendix \ref{homo_appendix}.

\noindent\textbf{Discussion.}
The consistent patterns between the original and poisoned prompted graphs demonstrate that backdoors against graph prompt tuning can be more evasive to homophily-based detectors \cite{unnoticeability_node_centric_homo}. These detectors might only be effective at identifying poisoned samples with significantly low homophily, which highlights the inherent vulnerability of graph prompt tuning. Additionally, the inverse pattern in the negative transfer scenario suggests that global-view homophily might explain why graph prompt tuning fails in such cases. We leave a detailed exploration of this phenomenon for future work.

\subsubsection{\textbf{Centroid Similarity}}
\begin{figure}[htbp]
    \centering
    \includegraphics[width=\linewidth]{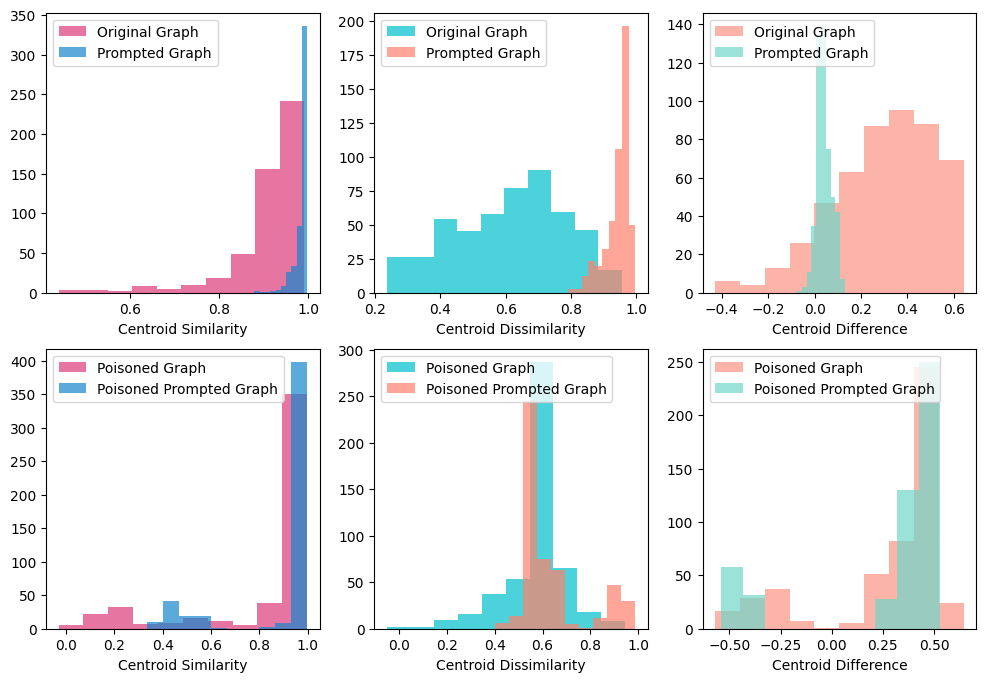}
    \caption{Centroid alignment (LHS), misalignment (Center) and difference (RHS) distributions of poisoned nodes. The pink, green, blue and orange have the same meanings as in Figure \ref{homo_bkd}. The salmon/cyan colors in the top denote the original graph/prompted graph while in the bottom denote the poisoned graph/poisoned prompted graph.}
    \label{sim_mis}
    \vspace{-4mm}
\end{figure}

We present the distributions of centroid alignment, misalignment, and difference in Figure \ref{sim_mis}.

\noindent\textbf{Takeaways.}
From Figure \ref{sim_mis}, we observe consistent patterns: after benign graph prompt tuning, both centroid alignment and misalignment increase. There are the same trend in graph backdoors. We utilize the centroid difference to discern their nuances. When tuning graph prompts either for original graphs or trigger-injected graphs, the centroid difference further decreases. This finding generalizes to other positive transfer scenarios, though contrary trends are observed in the negative transfer case. For more details, please refer to Appendix \ref{centroid_sim_appendix}.

\noindent\textbf{Discussion.}
In designing effective graph backdoor attacks, our goal is to achieve higher centroid alignment between poisoned samples and the target label's centroid, while minimizing the centroid misalignment between poisoned samples and their victim label's centroid, thus increasing the likelihood of misclassification. However, after graph prompt tuning, we observe a decrease in centroid difference, which contradicts our expectations. To address this challenge and enhance backdoor effectiveness, we incorporate this constraint into the final loss function, as outlined in the following section.

\subsubsection{\textbf{Embedding Space Visualization}}
We employ TSNE to project the embeddings of ego-networks into two dimensions, with poisoned nodes marked as stars. Due to space limits, we present them in Appendix \ref{emb_space_appendix}.

\noindent\textbf{Takeaways.}
From Figure \ref{emb_space}, before graph prompt tuning, the intra-distance of poisoned samples (without label flipping) is relatively large. After graph prompt tuning, this distance further decreases. Without graph prompt tuning, poisoned samples cluster together, remaining far from nodes with other labels and separated from nodes with the target label. With graph prompt tuning, the distance between poisoned samples and nodes with the target label further decreases, and the clean samples exhibit more compact patterns. These findings are consistent across other positive transfers. In the negative transfer case, all nodes, except for the poisoned samples in the poisoned prompted graph, begin to interlace again. For more details, please refer to Appendix \ref{emb_space_appendix}.



\begin{figure}[htbp]
    \vspace{-2mm}
    \centering
    \includegraphics[width=1\linewidth]{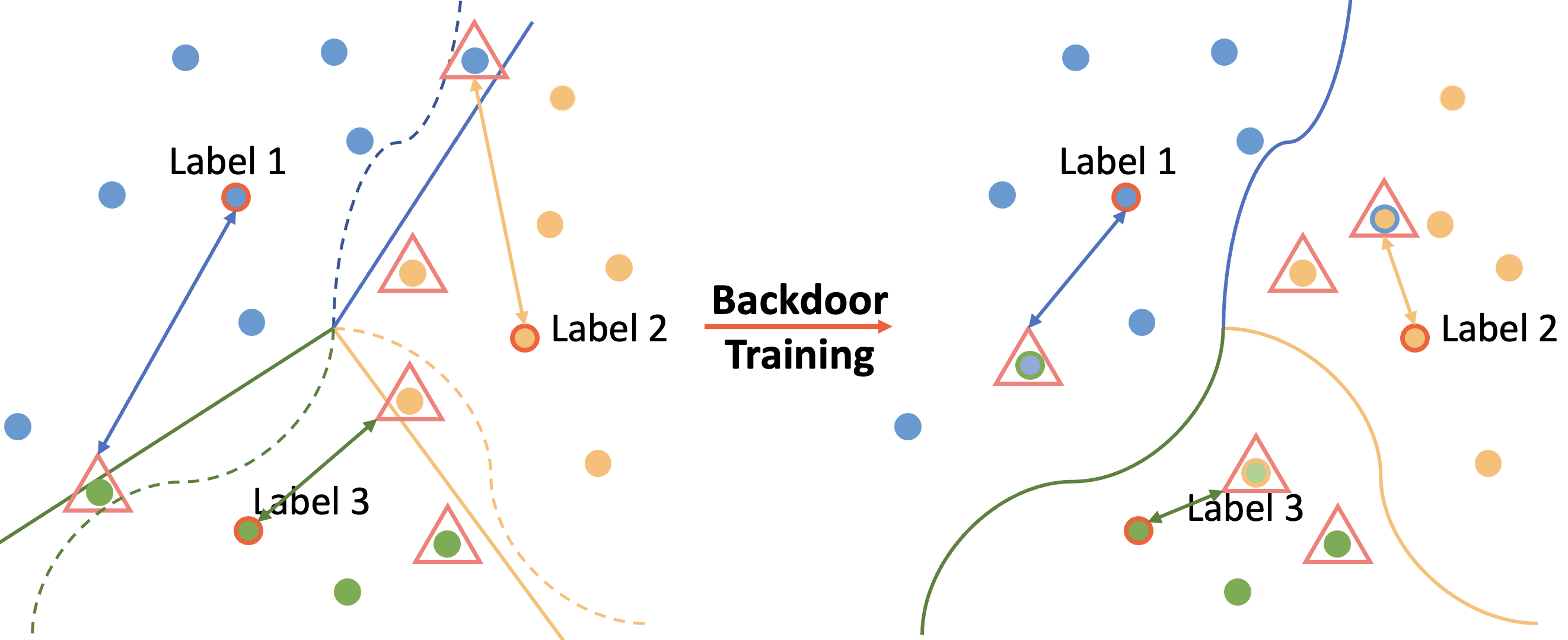}
    \caption{An illustrative example of selecting boundary nodes and pushing them closer to the centroid of the target label and farther away the centroid of the victim label.}
    \label{push_boundary}
    \vspace{-4mm}
\end{figure}

\noindent\textbf{Discussions.} If defenders can access intermediate outputs, e.g., node embeddings, they can project these into $2$-dimensions. This strategy allows them to trace decision boundaries and identify nodes that cluster together but remain isolated from other nodes in the same label group, marking them as highly suspicious poisoned samples. However, this detection method requires attackers to design more stealthy graph backdoors, pushing poisoned samples closer to the centroid of the target label and farther from the centroid of the victim label. This goal aligns with the aforementioned analysis of centroid similarity. A sketch map of this process is shown in Figure \ref{push_boundary}.

\vspace{-3mm}
\subsection{Problem Statement}
Given a downstream graph $\mathcal{G}=(A,X)$ with a label set $|Y|$, a pre-trained GNN $f_{pre}$, a graph prompt tuning model $f_{prm}$, a prompt generator $g_{\psi}$, a universal trigger $\delta$, the maximized trigger size $\tau$ and the attack budget $p$, the attacker aims to implant an effective and stealthy graph prompt backdoor as depicted in Equation (\ref{backdoor_express}):
\begin{equation} 
    s.t. |\delta|\leq \tau; |\mathcal{V}_p|\leq p
\end{equation}
\vspace{-2mm}
\section{Attack Framework Design}
\label{sec:framework}
\begin{figure}[h]
  \centering
  \includegraphics[width=\linewidth]{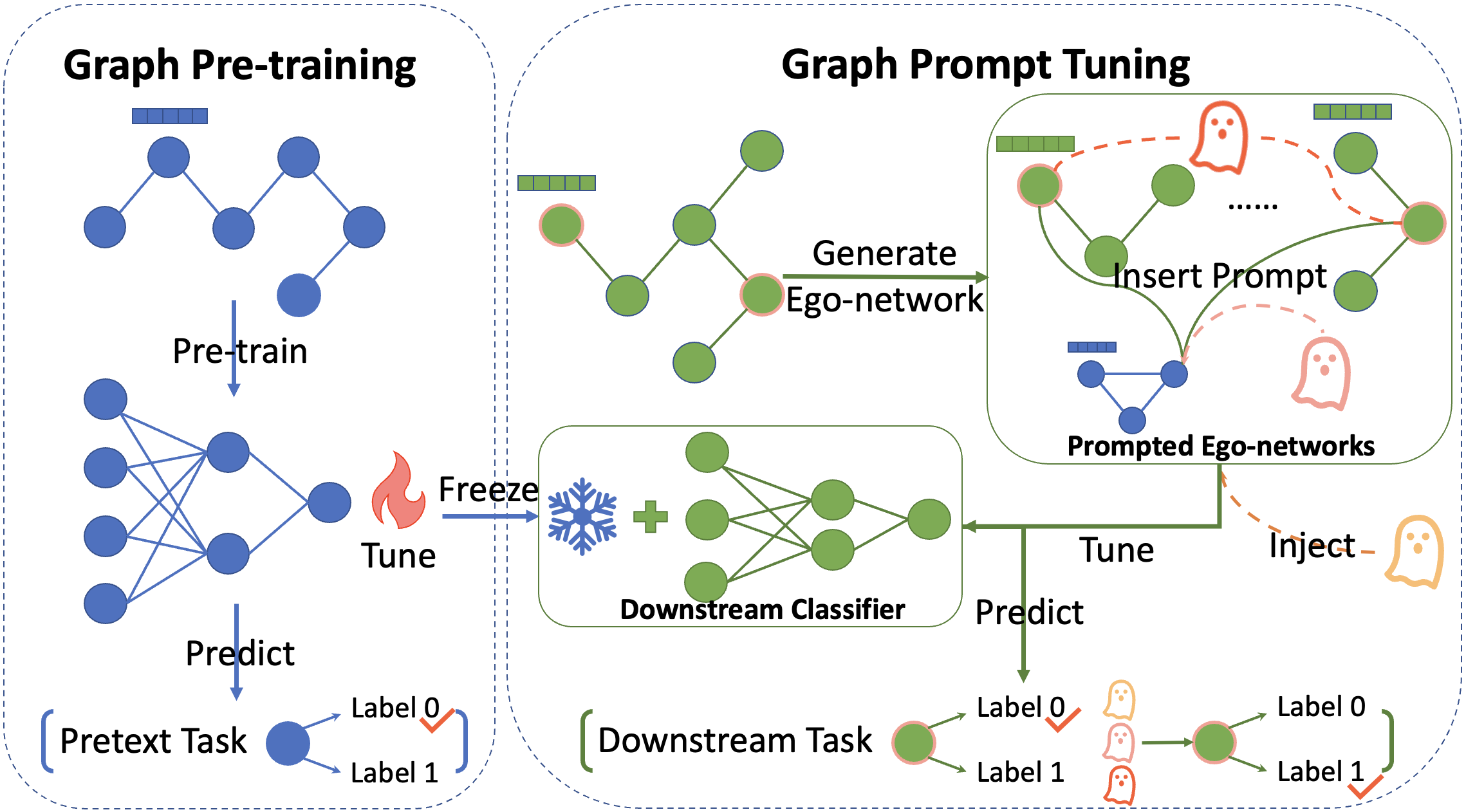}
  \caption{Krait pipeline Overview: the fire and snow marks denote tuning and freezing the pre-trained GNNs, and the red, pink, and yellow ghosts denote ``Invoke", ``Interact" and ``Modify" trigger generation methods, respectively.}
  \label{overview_krait}
  \vspace{-4mm}
\end{figure}

In this section, we present Krait, an effective and stealthy graph prompt backdoor to answer the third question. The overview of Krait is shown in Figure \ref{overview_krait}. Notably, graph pre-training is treated as a black-box and attackers cannot access pre-training graphs and associated graph pre-training process.

\vspace{-4mm}
\subsection{Threat Model}

\subsubsection{\textbf{Attacker's Goal}}
As formulated above, the attacker aims to manipulate the graph prompt tuning model such that poisoned nodes attached with triggers are misclassified as the target label, while maintaining the performance of clean nodes.

\subsubsection{\textbf{Attacker's Knowledge and Capabilities}}

We assume that attackers can only access downstream graphs but cannot access pre-training graphs or corresponding pre-training process. For downstream tasks, attackers are prohibited from modifying inner parameters or deactivating neurons. However, they can access intermediate outputs (e.g., node embeddings), provide customized loss functions for their desired tasks, and attach triggers to test nodes during the inference phase. They can interact with graph prompts through a specific user interface or by gaining partial control of prompt tuning. 
Notably, these setups are minimal to implement Krait, they also work under the white-box setting with full model control. Besides, interactive graph prompts are feasible, especially under a new trend of combining graph models with LLMs, where the latter respond to a wide range of hand-crafted prompts and can simultaneously learn multiple prompts \cite{multi_prompt,multi_prompt_recommendation}. Our attack setting is more stringent than the classical white-box settings, making it more practical and generalizable in real-world scenarios. We further extend it into the black-box setting.

\subsubsection{\textbf{Attack Scenarios}} Krait can be applied to: 

\noindent\textbf{Malicious users in trusted APIs or platforms}. Graph prompt tuning models can be integrated into APIs or deployed on platforms. However, users in these trusted environments can be malicious. Our attack setting is practical as many platforms such as Microsoft Azure and Google Vertex, support custom training, exposing the risks of manipulating intermediate outputs and modifying loss functions for malicious purposes. \par
\noindent\textbf{Malicious model providers}. Malicious model providers can gain full control over the training process, enabling them to implant more powerful graph backdoors. Based on practical constraints like capital and technology limitations, we adopt a conservative assumption about the adversary's capabilities. In reality, since only partial services are supported by outsourced parties, providers may be restricted to calling the interface for custom training or opting for cheaper APIs to fulfill their specific needs. Under these rigorous situations, they could modify users' customized graphs and leverage interactive prompting windows to launch graph backdoor attacks. \par
\noindent\textbf{Insecure collaborations among different departments or parties}. Due to disparate security and privacy restrictions, departments or business entities in the graph transfer learning pipeline may only offer interfaces or APIs for downstream custom training, with no access to the pre-training graphs and process. External attackers can bypass security checks to implant graph backdoors through interactive prompts. 

\vspace{-2mm}
\subsection{Framework Design}
Existing research on graph backdoor attacks face several key challenges: 1) generating random subgraphs or injecting triggers into randomly selected training nodes, ignoring rich information embedded in node features and topological structures to facilitate graph prompt backdoors \cite{random_backdoor}; 2) relying on computational-intensive methods to learn injection positions \cite{graph_backdoor, explain_backdoor}; 3) consuming large attack budgets (e.g., 10\% PR) to promise considerable attack effectiveness \cite{unnoticeable_graph_backdoor, contrastive_backdoor, motif_backdoor}; 4) cannot generate into more advanced attack types, such as all-to-one and all-to-all attacks. These gaps naturally raise the first research question: how can we identify the most vulnerable poisoned candidates with limited attack budgets across various attack types before model training? 

One spontaneous solution is to locate nodes that are easier to misclassify. Intuitively, they always lie near decision boundaries and are more sensitive to minimal perturbations. Thus, the first step is to identify these vulnerable boundary nodes.

\subsubsection{\textbf{Poisoned Candidate Selection}}
We design a new poisoned candidate selection metric to enhance the effectiveness of graph prompt backdoor attacks. 

\noindent\textbf{Label Non-uniformity.}
Label non-uniformity \cite{label_non_uni} is one of the pertinent metrics to identify boundary nodes based on label distribution, which is expressed as follows: 
\begin{equation}
    w(v)=\sum\limits_{y\in\mathcal{Y}}\bigg|\mu_{v}(y)-\frac{1}{|\mathcal{Y}|}\bigg|
\end{equation}
where $\mu_{v}(y)\in\left[0,1\right]$ denotes the soft-label prediction of node $v$ that belongs to label $y$. $|\mathcal{Y}|$ denotes the number of labels, and $\mathcal{Y}$ should be a finite label set and $\sum_{y\in\mathcal{Y}}\mu_{v}(y)=1$. The smaller value of $w(v)$ indicates that the node $v$ is closer to class boundaries, thus harder to correctly classify. However, this metric can only be obtained after GNN training, which inspires us to design a model-agnostic selection rule during the pre-processing phase. The reasons are as follows: 1) Attackers prefer not to heavily query the victim models, as in real-world scenarios, query budgets are usually limited, and frequent queries are more likely to be monitored and detected by defenders; 2) Due to strong security and privacy concerns, many models and systems only support hard-label predictions, making it difficult to leverage post-training metrics which heavily depend on soft-label predictions; 3) Attackers may consider launching model stealing attacks to construct surrogate models. However, these attacks are time-consuming and require frequent queries. Additionally, the quality of soft-label predictions obtained through model stealing is highly dependent on the effectiveness of surrogate models; 4) Some existing studies utilize explanation-based models \cite{explain_backdoor} to select poisoned candidates, which involve high computational costs for perturbation and sampling, especially in large-scale graphs.  

\begin{figure}[htbp]
    \vspace{-2mm}
    \centering
    \subfloat[\centering{High homophily \& high LDN}]{  \includegraphics[width=0.315\linewidth]{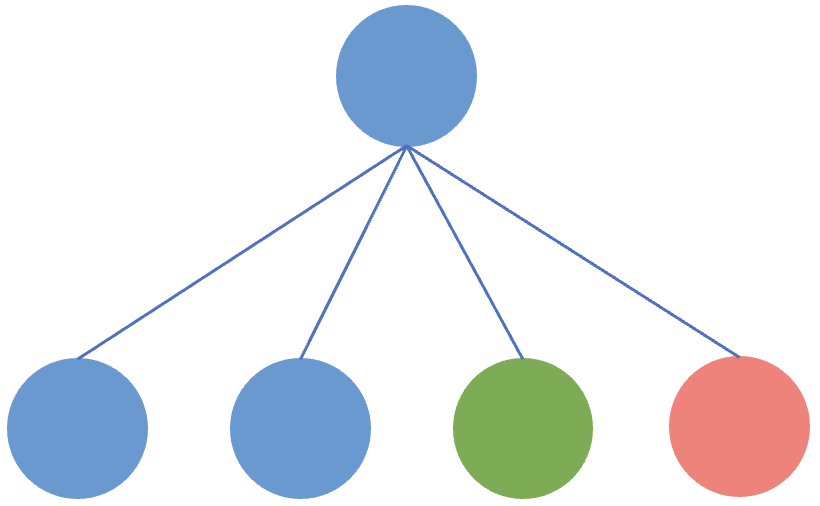}}
    \subfloat[\centering{Low homophily \& high LDN}]{   \includegraphics[width=0.315\linewidth]{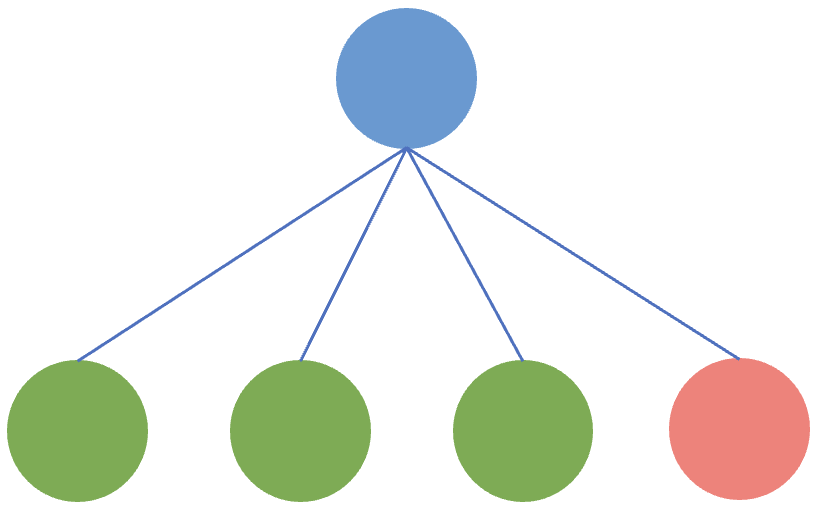}}
    \subfloat[\centering{Low homophily \& low LDN}]{  \includegraphics[width=0.315\linewidth]{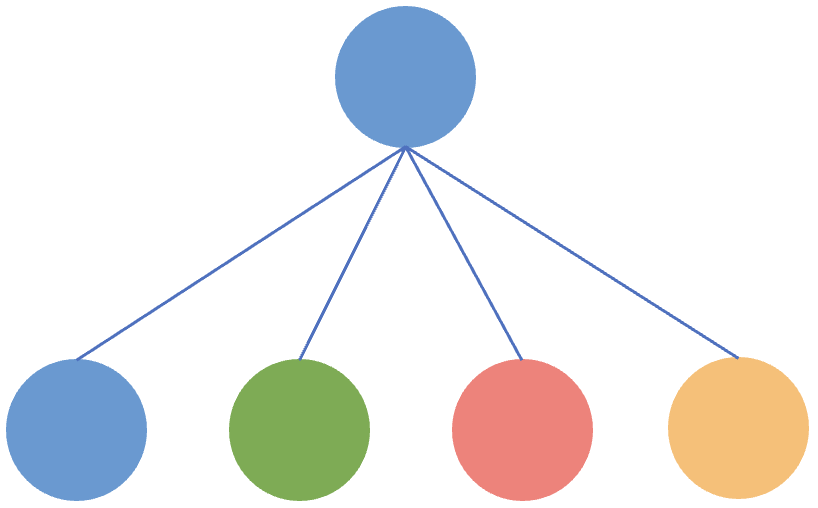}}
    \vspace{-4mm}
    \caption{Toy examples of label non-uniformity homophily.}
    \label{lnh_4_cases}
    \vspace{-6mm}
\end{figure}

\noindent\textbf{Label Non-uniformity Homophily.} Since most GNNs intrinsically follow homophily assumptions, homophily can be an effective pre-training metric for assessing misclassification risks. Nevertheless, as indicated in previous works \cite{homo_class1,homo_class}, GNNs still perform well in the absence of homophily. Additionally, traditional label-based homophily metrics only consider the count of neighbors with the same label rather than the neighbors' label distribution. Similarly, node-centric homophily \cite{unnoticeability_node_centric_homo} measures node feature similarity but ignores label distribution. To address these gaps, we propose a new metric called label non-uniformity homophily (LNH), which aligns homophily with label distribution non-uniformity (LDN). Since homophily can be generated from the label distribution, when homophily is high, the corresponding LDN is high as well. However, when homophily is low, LDN is less predictable since the majority of neighbors may be from the same label group but different from the targeted node's label, leading to high LDN. We present toy examples in Figure \ref{lnh_4_cases} to illustrate these concepts. Intuitively, homophily reflects how a targeted node benefits from its neighbors during GNN training, and LDN indicates the degree of updating new information from nodes with different labels. For backdoor attacks, we specifically focus on nodes that exhibit low homophily and high LDN. These nodes are more likely to be misclassified as the specific label since most of their neighbors share this label. To align the changing directions of homophily and LDN, for each node $v\in\mathcal{V}$, we define LNH as:
\begin{equation}
    h_v=\left(1-\frac{1}{|\mathcal{N}(v)|}\sum\limits_{u\in\mathcal{N}(v)}\mathbb{I}(y_u=y_v)\right)\sum\limits_{y\in\mathcal{Y}}\bigg|\tau(\mathcal{N}(v),y)-\frac{1}{|\mathcal{Y}|}\bigg|
\label{LNH}
\end{equation}
where $\tau(\mathcal{N}(i),y)=\frac{1}{|\mathcal{N}(i)|}\sum_{u\in\mathcal{N}(i)}\mathbb{I}(y_u=y)$, $\tau(\mathcal{N}(i),y)\in [0,1]$, $\sum_{y\in\mathcal{Y}}\tau(\mathcal{N}(i),y)=1$. 

Attackers can compute LNH for each node with the victim label in the customized graph, rank them in descending order of LNH values, and select those with larger LNH values as potential poisoned candidates. They can combine with the degree filter to retain poisoned samples with smaller degrees. This is because injecting triggers into nodes with larger degrees will largely degrade the GNNs' performance \cite{unnoticeable_graph_backdoor}. The algorithm encapsulating these steps is detailed in Appendix \ref{algo1}.

\subsubsection{\textbf{Attack types}}
We consider three attack types in our paper: one-to-one, all-to-one, and all-to-all attacks \cite{first_work,all_to_one_all}.

\noindent\textbf{One-to-one attack.} The classical approach involves randomly selecting any label or simply choosing the first label as the target label \cite{random_backdoor,graph_backdoor,contrastive_backdoor,explain_backdoor,motif_backdoor,unnoticeable_graph_backdoor}. We argue that this strategy is impractical since attackers may have their own `malicious preference'. For instance, they tend to inject triggers into `harmful content' and manipulate it into `harmless' in recommendation systems, or alter `default' into `qualified applicants' in financial crediting practices. Since these hazardous contents merely occupy a tiny fraction,
unlike the settings in Section \ref{sec:prelim_exp}, we assume that the target label is the label with the largest number of nodes, while the victim label is the one with the smallest number of nodes. Attackers can utilize LNH to select poisoned candidates closer to the decision boundary between the victim and the target label.

\noindent\textbf{All-to-one attack.} Attackers can choose the same target label as in one-to-one attacks, and simultaneously misclassify nodes with the remaining labels into the target one. In high-dimensional space, there may exist a sub-decision boundary (even if not that clear) between the target label and the rest. Attackers can select a group of nodes from each remaining label as poisoned candidates.

\noindent\textbf{All-to-all attack.} We follow the same strategy in Doan et al. \cite{all_to_one_all} to choose attack pairs based on the victim label $y$ as:
\begin{equation}
    Y_t = (y+1) mod |\mathcal{Y}|
\end{equation}
\vspace{-2mm}
\subsubsection{\textbf{Trigger Generation and Injection}}
Graph prompts can be seen as additional subgraphs, which in essence are the same as triggers \cite{unnoticeable_graph_backdoor}. As depicted in Section \ref{sec:prelim_exp}, attackers can exploit this inherent vulnerability to disguise triggers as benign prompts. Hence, we adopt the ``using somebody’s spear to attack his shield" strategy to generate and attach universal triggers $\delta$ for poisoned candidates $\mathcal{V}_{p}$ through benign graph prompt tuning $g_{\psi'}$, which is formulated as $g_{\psi'}(\mathcal{G}_{s(v)})=\mathcal{G}_{s(v)}\oplus\delta$ for $v \in \mathcal{V}_{p}$. The optimization can be expressed as: 
\begin{equation}
   \min\limits_{\psi',\theta}\mathcal{L}_{gen}=\sum\limits_{v\in\mathcal{V}_{p}}l(f_{gen}(g_{\psi'}(\mathcal{G}_{s(v)}), \theta_{gen}), Y_t)
\label{gen}
\end{equation}
where $f_{gen}$ and $\theta_{gen}$ are defined similarly as before. But Equation (\ref{prm}) requires flipping the label of poisoned candidates and generating graph prompts as triggers for these candidates, while Equation (\ref{gen}) generates benign graph prompts for all training nodes. As this process involves two prompt generation steps, we propose the following three trigger generation methods based on their sequence. 

\noindent\textbf{Double Prompt-invocation Method (Invoke).} Attackers can first utilize open-sourced graph prompt tuning models $g_{\psi_o}$ (e.g. SOTA models in the ProG package) to generate graph prompts for their meticulously designed warm-up samples as triggers $\delta_{fix}$, using the same optimization process in Equation (\ref{gen}). In the context of all-to-one attacks, flipping all victim labels to the target label means that the entire graph is labeled the same, yielding ineffective triggers since the lazy downstream classifiers can directly predict all samples as the same label. Therefore, attackers can poison the majority, e.g, $90\%$ of training samples with the victim label as a poisoned warm-up. Please note that this warm-up is feasible given that attackers have full control over the open-sourced models on their end. Finally, attackers can attach the fixed triggers $\delta_{fix}$ to poisoned candidates $\mathcal{V}_{p}$, and feed them along with clean samples into the regular graph prompt tuning process. The whole optimization process is depicted in Equation (\ref{backdoor_express}). Notably, attackers can exploit benign graph prompt tuning to generate poisoned prompts as triggers, making the attack more covert. However, this method requires invoking models twice, resulting in higher computational overhead. Luckily, since graph prompt tuning can adapt pre-trained GNNs to downstream tasks within a few steps, the overall costs are acceptable.

\noindent\textbf{Multiple Prompt-interaction Method (Interact).} Rather than invoking the model twice, attackers can interact with the graph prompt tuning model in a sequential manner. They can manipulate the victim model to generate graph prompts $\delta$ for the selected poisoned candidates $\mathcal{V}_{p}$ using Equation (\ref{gen}), thereby updating $\mathcal{G}_{s(v)}$ to $\mathcal{G}_{s(v)}'=\mathcal{G}_{s(v)}\oplus\delta$; and then utilize benign graph prompt tuning to create ``benign" graph prompts for the poisoned graph. This method relies on benign graph prompt tuning to craft poisoned prompts as triggers, allowing attackers to learn both the universal triggers and the ``benign" graph prompts within a single training process as follows: 
\begin{equation} 
\begin{split}
    \min\limits_{\theta_{tri}}\mathcal{L}_{tri}&= \sum\limits_{v\in\mathcal{V}_{c}}l(f_{prm}(g_{\psi}(\mathcal{G}_{s(v)}), \theta), Y_v)+ \\    &\phantom{=\;\;}\sum\limits_{u\in\mathcal{V}_{p}}l(f_{prm}(g_{\psi}(\mathcal{G}_{s(u)}'), \theta), Y_t)
\end{split}
\label{backdoor_interact}
\end{equation}

\vspace{-3mm}
\noindent\textbf{Prompt Modification Method (Modify).} Instead of interacting with graph prompts, attackers can directly modify existing graph prompts $\mathcal{G}_{\Delta}$ for selected poisoned candidates $\mathcal{V}_{p}$. Similar to graph injection attacks \cite{unnoticeability_node_centric_homo}, attackers can inject triggers of the same size as graph prompts to the prompted graph. These triggers can be disguised as benign graph prompts and simultaneously tuned along with the real benign graph prompts, which enhances their stealthiness. This method resembles the ``Interact" method but exploits graph prompts in reverse order: 
\begin{equation} 
\begin{split}
    \min\limits_{\theta_{tri}}\mathcal{L}_{tri}&= \sum\limits_{v\in\mathcal{V}_{c}}l(f_{prm}(g_{\psi}(\mathcal{G}_{s(v)}), \theta), Y_v)+ \\    &\phantom{=\;\;}\sum\limits_{u\in\mathcal{V}_{p}}l(f_{prm}(g_{\psi'}(g_{\psi}(\mathcal{G}_{s(u)})), \theta), Y_t)
\end{split}
\label{backdoor_modify}
\end{equation}
\vspace{-3mm}

\subsubsection{\textbf{Loss Functions}}
The core loss function is depicted in Equation (\ref{backdoor_express}). Here we introduce the aforementioned centroid difference as the constraint. The final loss function can be expressed as:
\begin{equation} 
    \min\limits_{\theta_{tri}}\mathcal{L}_{bkd} = \mathcal{L}_{tri}+\frac{\alpha}{|\mathcal{V}_{p}|}\sum\limits_{j\in\mathcal{V}_{p}}[-\mathcal{CF}_{j}+\beta]_{+}
\end{equation}
where we can treat the third term as a special form of triplet loss since the centroid difference of node $j$ can be seen as the distance between node embedding $Z_j$ and negative sample $N_j$ minus the distance between $Z_j$ and positive sample $P_j$. $Z_j$ is the anchor, $P_j$ and $N_j$ are the centroids of the target label and the victim label, respectively. $\beta$ is the margin, the default as $1$, and $\alpha$ denotes the coefficient of triplet loss. The Krait algorithm is detailed in the Appendix \ref{algo2}.
\vspace{-2mm}
\subsection{Extension: Black-box Backdoor against Graph Prompts}
\label{black_box}
We extend Krait into the black-box setting. We assume that attackers cannot design customized loss functions, but can access downstream graphs and interact with vulnerable graph prompts. Attackers are allowed to obtain some side information as prior knowledge, such as the value of $k$ in $k$-hop ego-network constructions, which is practical as this information can be acquired in research papers, user manuals, open-sourced tutorials, etc. Existing black-box backdoor attacks \cite{b3_black_box_backdoor} predominantly utilize model stealing attacks to develop surrogate models, implant backdoors to these copycats, and finally release them to the public. However, these methods encounter problems such as high computational costs, unstable quality of surrogate models, and the risk that users might not be easily deceived by such copycats.

To save space, we solely discuss the technical details that are different from the aforementioned attacks.

\noindent\textbf{Attack Pipeline.}
Under the black-box setting, a straightforward way is to leverage three trigger generation methods without requiring customized loss functions. Moreover, we consider a more stringent setting where multiple prompt interactions are not feasible. To tackle these problems, similar to the ``Invoke" method, attackers can employ open-sourced graph prompt tuning models to generate universal graph prompts as triggers for warm-up poisoned samples. Subsequently, they can attach these triggers to poisoned candidates and feed them, along with clean samples, into a benign graph prompt tuning.
\vspace{-2mm}
\section{Experimental Validation}

\label{sec:exp}
In this section, we conduct empirical experiments based on different types of attacks and trigger generation methods to investigate the following key research questions:

\noindent\textbf{RQ1}: How effective/stealthy is Krait for graph backdoors?

\noindent\textbf{RQ2}: How does Krait perform under diverse combinations of attack types and trigger generation methods? Can it flexibly adapt to different GNN backbones?

\noindent\textbf{RQ3}: What factors can affect the performance of Krait?

\noindent\textbf{RQ4}: Can Krait be extended to the black-box setting?

\vspace{-3mm}
\subsection{Setups}

\noindent\textbf{Datasets.}
We use the same datasets described in Section \ref{sec:prelim_exp}.

\noindent\textbf{Model.}
We set All-in-One \cite{all_in_one} as the victim graph prompt tuning model, using the Graph Transformer (GT) \cite{graph_transformer} as the GNN backbone and GraphCL \cite{graphcl} for pre-training. We consider the multi-class classification setting.

\noindent\textbf{Baselines.}
To the best of our knowledge, Krait is the first graph prompt backdoor, there is no existing method to compare. Therefore, we use random selection and LNH selection respectively to choose poisoned candidates for each type of attack under diverse trigger generation methods as our baselines.

\noindent\textbf{Implementations.}
We follow the default parameter settings in All-in-One \cite{all_in_one}, please refer to more implementation details in Appendix \ref{imp_details}. Besides, we set adaptation steps as $10$, and the trigger size is the same as the graph prompt size, which is $10$. We utilize all nodes in each graph dataset instead of sampling a specific number of nodes for training and testing. We $1$:$1$ split the testing samples, half are clean samples and the rest are attached with triggers. And the coefficient $\alpha=10$. Notably, we poison 2\% of training nodes with the victim label for all experiments. The PRs relative to the entire training graph range from 0.15\% to 2\% across all three attack types. Since our attack target is mostly the label with the least nodes, the poisoning rates further decrease. We report the average results of $5$ trials. All experiments are conducted on a $64$-bit machine with $4$ Nvidia A$100$ GPUs. The code is \href{https://github.com/yings0930/Krait.git}{provided}.

\noindent\textbf{Metrics.}
For benign models, we use accuracy (ACC), F1 score, and AUC to evaluate the performance of node classification. For graph backdoors, to evaluate attack effectiveness, we employ 1) Attack Success Rate (ASR) to quantify how many trigger-injected samples are misclassified to the target label; 2) average misclassification confidence (AMC) \cite{graph_backdoor} to compute the average confidence score assigned to the target label for the successful graph backdoors. To evaluate attack stealthiness, we use 3) poisoning rate (PR) to represent attack budgets; 4) clean accuracy (CA) to measure the classification accuracy of clean samples after backdoor training; 5) average homophily difference (AHD) to gauge the local-subgraph homophily differences between clean samples and trigger-injected samples.

\vspace{-2.5mm}
\subsection{Attack Performance}
The performance of benign graph prompt tuning is shown in Table \ref{benign_perf} in Appendix \ref{benign_perf_app}. We observe that transfers from different domains are more unstable, and generally lead to worse performance than within-domain transfers, which fits our intuition. Since there exists up to $4\times4\times3\times3$ cases, based on Table \ref{benign_perf}, we solely pick up $3$ representative cases, namely, CiteSeer to Cora (within-domain, positive, small-to-small transfer), Physics to Computers (out-of-domain, relatively negative, and big-to-big transfer) and Physics to CiteSeer (out-of-domain, positive, and big-to-small transfer) to conduct corresponding experiments.
\vspace{-2mm}
\begin{table*}[htbp]
\caption{Performance of Krait under the white-box setting. Cite2Cora: CiteSeer to Cora transfer; Phy2Cite: Physics to CiteSeer transfer; Phy2Comp: Physics to Computers transfer. Random and LNH denote poison node selection strategies, and Full denotes Krait. We bold the best performance under different combinations of attack types and trigger generation methods.}
\resizebox{\linewidth}{!}{
\begin{tabular}{c|c|c|cccc|cccc|cccc}
\toprule
\textbf{Transfer}                      & \multirow{3}{*}{\textbf{Method}}   & \multirow{3}{*}{\textbf{Model}} & \multicolumn{4}{c|}{\textbf{One-to-One}}                                                            & \multicolumn{4}{c|}{\textbf{All-to-One}}                                                           & \multicolumn{4}{c}{\textbf{All-to-All}}                                                            \\ \cline{1-1} \cline{4-15} 
(Evaluation)                       &                           &                        & \multicolumn{2}{c|}{\textbf{Effectiveness}} & \multicolumn{2}{c|}{\textbf{Stealthiness}}       & \multicolumn{2}{c|}{\textbf{Effectiveness}} & \multicolumn{2}{c|}{\textbf{Stealthiness}}      & \multicolumn{2}{c|}{\textbf{Effectiveness}} & \multicolumn{2}{c}{\textbf{Stealthiness}}       \\ \cline{1-1} \cline{4-15} 
(Metrics)                          &                           &                        & ASR            & \multicolumn{1}{c|}{AMC} & CA                         & AHD            & ASR             & \multicolumn{1}{c|}{AMC} & CA                       & AHD          & ASR             & \multicolumn{1}{c|}{AMC} & CA                         & AHD          \\ \hline
\multirow{9}{*}{\textbf{Cite2Cora}}     & \multirow{3}{*}{\textbf{Invoke}}   
& \textbf{Random}                 & 94.22               & 93.12                         & 86.89          & \textbf{0.18}                                    & 29.10               & 91.49                         & 85.74          & 4.97                                             & 4.23               & 78.19                         & 85.95           & 9.54             \\
                                   &                           
& \textbf{LNH}                    & 95.56               & 93.20                        & \textbf{86.95}  & 0.19                                             & 38.68               & 90.97                        & 85.39           & 5.02                                             & 4.05               & 77.95                         & 85.63           & \textbf{9.53}             \\
                                   &                           
& \textbf{Full}                  & \textbf{100.00}       & \textbf{100.00}             & 86.75              & 3.02                                         & \textbf{100.00}       & \textbf{100.00}             & \textbf{85.83}    &\textbf{3.24}                                  & \textbf{6.67}         & \textbf{82.60}              & \textbf{86.10}     & 10.43             \\ \cline{2-3}
                                       & \multirow{3}{*}{\textbf{Interact}} 
& \textbf{Random}                 & 42.22 & 86.68           & 86.89  & 0.91 
                                 & 99.91 & 99.75            & 85.66  & \textbf{3.65} 
                                 & \textbf{42.44}  & \textbf{93.96}            & 85.74   & 6.19 \\
                                   &                           
& \textbf{LNH}                    & 56.00 & 89.37           & \textbf{86.95}  & \textbf{0.82} 
                                & 95.51  & 98.86            & 85.68  & 3.71 
                                & 38.38               & 90.53                         & 85.83     & 6.19             \\
                                   &                           
& \textbf{Full}                  & \textbf{62.22}         & \textbf{91.77}               & 86.72      & 1.29                                               & \textbf{100.00}               & \textbf{100.00}       & \textbf{85.86}    & 4.05                                           & 31.78      & 89.51         & \textbf{86.13}                 & \textbf{5.86}             \\ \cline{2-3}
                                   & \multirow{3}{*}{\textbf{Modify}}   
& \textbf{Random}                 & 43.56               & 85.88                         & 86.90         & \textbf{0.88}                                 & \textbf{100.00}               & 99.84              & 85.66               & \textbf{3.60}                               & \textbf{42.61}               & \textbf{93.30}              & 85.83          & 6.38             \\
                                   &                           
& \textbf{LNH}                    & 54.67               & 89.64                & \textbf{86.98}                  & 0.93                                     & 96.53               & 98.83                & 85.74                          & 3.79                                        & 38.68               & 89.56              & 85.83                          & 6.13             \\
                                   &                           
& \textbf{Full}                  & \textbf{59.56}          & \textbf{91.83}             & 86.69       & 1.29                                               & \textbf{100.00}               & \textbf{100.00}        & \textbf{85.95}    & 4.07                                      & 31.89               & 89.05            & \textbf{86.27}              & \textbf{5.80}            \\ \hline
\multirow{9}{*}{\textbf{Phy2Cite}}  & \multirow{3}{*}{\textbf{Invoke}}   & \textbf{Random}                 & 83.64               & 98.76                         & 81.56                          & \textbf{9.61}              & 17.14               & 95.05                         & \textbf{81.47}                            & \textbf{3.13}             & 8.60                           & \textbf{94.66}              & \textbf{81.11}                                & 3.13        \\
                                   &                           & \textbf{LNH}                    & 78.79               & 97.51                         & 81.78                        & 9.64              & 17.89               & 96.40                         & 80.63                            & 3.15             & 8.46               & 93.19                         & 79.62                            & 3.31             \\
                                   &                           & \textbf{Full}                  & \textbf{100.00}               & \textbf{100.00}                         & \textbf{82.03}                             & 11.01              & \textbf{100.00}               & \textbf{100.00}                         & 79.90                             & 3.97             & \textbf{8.81}               & 92.17                         & 80.34                           & \textbf{2.65}             \\ \cline{2-3}
                                   & \multirow{3}{*}{\textbf{Interact}} & \textbf{Random}                 & 23.64               & 65.49                         & 81.53                             & \textbf{9.74}              & 97.98               & 99.81                         & 81.15                          & \textbf{3.29}             & 15.87               & 94.00                         & 80.87                           & 3.22             \\
                                   &                           & \textbf{LNH}                    & 28.79               & 89.50                         & 80.93                          & 10.03              & 82.79               & 98.47                         & 80.00                            & 3.39             & 14.45               & 93.84                         & \textbf{80.91}                            & \textbf{3.13}             \\
                                   &                           & \textbf{Full}                  & \textbf{31.82}               & \textbf{93.12}                         & \textbf{81.84}                            & 10.09              & \textbf{99.93}               & \textbf{100.00}                         & \textbf{81.27}                             & 4.02             & \textbf{17.29}               & \textbf{95.58}                         & 79.50                             & \textbf{3.13}             \\ \cline{2-3}
                                   & \multirow{3}{*}{\textbf{Modify}}   & \textbf{Random}                 & 21.21               & 89.74                         & \textbf{81.45}                             & \textbf{9.70}              & 97.67               & 99.85                         & \textbf{81.30}                             & \textbf{3.35}             & 14.17               & 95.01                         & \textbf{81.01}                             & 3.24             \\
                                   &                           & \textbf{LNH}                    & 32.42               & 89.45                         & 80.80                             & 9.99              & 83.12               & 98.45                         & 80.02                            & 3.41             & 15.80               & 94.54                         & 80.36                             & 3.51             \\
                                   &                           & \textbf{Full}                  & \textbf{43.33}               & \textbf{92.28}                         & 81.21                             & 10.02              & \textbf{99.98}               & \textbf{100.00}                        & 81.03                             & 4.01             & \textbf{16.11}               & \textbf{96.25}                         & 79.66                            & \textbf{3.04}             \\ \hline
\multirow{9}{*}{\textbf{Phy2Comp}} & \multirow{3}{*}{\textbf{Invoke}}   & \textbf{Random}                 & 53.51               & 95.54                         & 62.21                            & -5.19              & 74.42               & 98.00                         & \textbf{62.93}                            & \textbf{3.08}             & 4.88               & 93.17                         & 61.02                            & 3.10             \\
                                   &                           & \textbf{LNH}                    & 43.24               & 94.62                         & 62.25                             & \textbf{-5.30}              & 73.56               & 98.30                         & 60.88                            & 3.12             & 3.94               & 93.97                         & 61.67                             & \textbf{3.02}             \\
                                   &                           & \textbf{Full}                  & \textbf{99.46}               & \textbf{100.00}                         & \textbf{63.31}                             & 4.73             & \textbf{100.00}               & \textbf{100.00}                         & 62.74                             & 3.64             & \textbf{25.30}               & \textbf{98.31}                         & \textbf{62.70}                              & 3.54             \\ \cline{2-3}
                                   & \multirow{3}{*}{\textbf{Interact}} & \textbf{Random}                 & 25.41               & \textbf{93.02}                         & 60.85                  & -5.25              & 98.16               & 99.77                         & 61.20                            & \textbf{3.55}             & 5.67               & 95.16                         & 61.50                           & 2.97            \\
                                   &                           & \textbf{LNH}                    & 24.05               & 91.85                         & \textbf{62.16}                             & \textbf{-5.27}              & 98.14               & 99.78                         & 60.28                             & 3.56             & 7.52               & 93.37                         & 61.62                             & \textbf{2.73}             \\
                                   &                           & \textbf{Full}                  & \textbf{31.08}               & 91.61                         & 61.70                         & -5.19              & \textbf{99.82}               & \textbf{100.00}                         & \textbf{63.75}                             & 3.90             & \textbf{18.37}               & \textbf{99.06}                         & \textbf{62.64}                           & 3.11             \\ \cline{2-3}
                                   & \multirow{3}{*}{\textbf{Modify}}   & \textbf{Random}                 & 34.86               & \textbf{94.35}                         & 62.15                            & -5.23              & 98.76               & 99.84                         & 60.19                            & \textbf{3.51}             & 6.13               & 96.51                         & 61.17                             & 3.15             \\
                                   &                           & \textbf{LNH}                    & 26.49               & 92.26                         & \textbf{62.28}                             & \textbf{-5.28}              & 98.26               & 99.79                         & 60.01                             & \textbf{3.51}             & 10.50               & 95.72                      & \textbf{61.76}                             & \textbf{2.55}             \\
                                   &                           & \textbf{Full}                  & \textbf{38.11}               & 90.45                         & 61.69                            & -5.20              & \textbf{99.56}               & \textbf{99.98}                         & \textbf{62.89}                             & 3.94             & \textbf{17.75}               & \textbf{98.95}                         & 59.87                             & 3.16             \\ \hline
\end{tabular}
}
\label{Krait_perf}
\vspace{-6mm}
\end{table*}
To address RQ1 and RQ2, we compare Krait with baselines under various attack types and trigger generation methods across three transfer cases.

\noindent\textbf{Effectiveness and Efficiency.}
From Table \ref{Krait_perf}, we observe that Krait consistently excels in terms of ASR and AMC, particularly in big-to-big and big-to-small transfers. In almost all attacks, Krait accomplishes higher attack effectiveness compared to the baselines. Notably, solely employing a label non-uniformity homophily strategy without constraints serves as an ablation study, where Krait surpasses or achieves considerable attack performances than random triggers in most attack combinations. Since Krait identifies trigger positions before any training process, it is much more computationally efficient compared to existing graph backdoor attacks that leverage the bi-level optimization to select poisoned candidates during model training. The time complexity of this process is $O(|V_{vic}|*d_{max})+O(rank)$, where $|V_{vic}|$ is the number of nodes with the victim label, $d_{max}$ is the maximum degree of these nodes, `rank' denotes any ranking method like the Quickselect that can be linear time \cite{quick_select}. Therefore, the efficient time complexity can be $O(|V_{vic}|*(d_{max}+1))$, which is comparable to $O(|V_{vic}|*d_{max})$. In our experiments, it only takes around 0.04-0.82 seconds to choose nodes from the victim label, compute their LNH, apply the degree filter, and rank their values across different-scale datasets.

\noindent\textbf{Stealthiness.}
As the PR ranges from $0.15\%$ to $2\%$ of the entire training graph across three attack types, it is significantly lower than the $10\%$ standard for graph backdoors. In one-to-one attacks for small-scale transfers, poisoned samples contain only two nodes; even in more complex scenarios, Krait only poisons $22$ nodes but achieves $100$\% ASR. Besides, compared to the benign classification accuracy of all test samples reported in Table \ref{benign_perf}, Krait achieves comparable clean accuracy or even surpasses the benign results in all attack types, demonstrating its stealthiness. Regarding statistical properties, the overall homophily changes between clean samples and poisoned nodes are slight (please note that AHD is expressed in percentage), which indicates that Krait effectively evades detection against homophily and node similarity-based defenders \cite{unnoticeability_node_centric_homo}. Moreover, Krait utilizes graph prompt tuning itself to generate graph prompts as triggers based on node similarities, making it more difficult to detect by node-similarity defenders. We further discuss the details in Subsection \ref{discuss_def}.

\noindent\textbf{Practicability.}
As shown in Table \ref{Krait_perf}, Krait exhibits superior performance with ASRs of $99.56\%-100\%$ in all-to-one attacks across all trigger generation methods. The ``Invoke" method excels in one-to-one attacks, achieving ASRs between $99.46\%$ and $100\%$, while the ``Interact" method stands out in all-to-all attacks, with ASRs ranging from $6.67\%$ to $31.89\%$. The relatively lower performance in all-to-all attacks is reasonable since Krait employs a universal trigger, differing from the specific triggers used in the computer vision (CV) domain \cite{first_work,all_to_one_all} that require learning distinct triggers for each attack victim and target pair. {We provide an in-depth bad-case analysis in Appendix \ref{bad_effective}. In real-world applications, attackers can consider their resources, capabilities, preferences, and involved stages to select the most suitable trigger-generation method.

\noindent\textbf{Flexibility.}
We select three combinations of attack types and trigger generation methods, namely, ``Invoke" for one-to-one attacks, ``Interact" for all-to-one attacks, and ``Modify" for all-to-all attacks in the CiteSeer to Cora transfer case. Except for GT \cite{graph_transformer}, we additionally employ the graph convolutional network (GCN) \cite{GCN} and graph attention network (GAT) \cite{GAT} as GNN backbones. Our experimental results show that Krait maintains superior performance in terms of attack effectiveness and stealthiness across different GNN backbones. Specifically, Krait embedded with GT as the backbone excels in most attack combinations. And Krait embedded with different GNN backbones all achieve nearly $100\%$ ASRs in all-to-one attacks that utilize the "Interact" trigger generation method. Please refer to Table \ref{backbone_Krait} in Appendix \ref{flex_perf} for more details. 

\vspace{-4mm}
\subsection{Parameter Studies} To answer RQ$3$, we conduct parameter studies on trigger size, PR, epochs, and constraint coefficient, respectively.

\noindent\textbf{Trigger size.} We vary trigger size as $[1,5,10,15,20]$. From Figure \ref{tri_size}, Krait consistently surpasses the baselines as trigger size increases. Notably, even trigger size is set to $1$, Krait still achieves $100\%$ ASRs in one-to-one and all-to-one attacks. However, larger trigger sizes in all-to-all attacks may result in ASR drops. We speculate that since Krait generates a universal trigger for all given attack pairs, it can continuously encode general malicious knowledge to facilitate graph backdoors before the trigger size reaches an ``optimal" threshold. As trigger size grows, Krait may encode redundant information, leading to ASR drops. Sun et al. \cite{all_in_one} report similar observations when increasing the size of the prompt token.
\vspace{-2mm}
\begin{figure}[htbp]
  \centering
  \includegraphics[width=\linewidth]{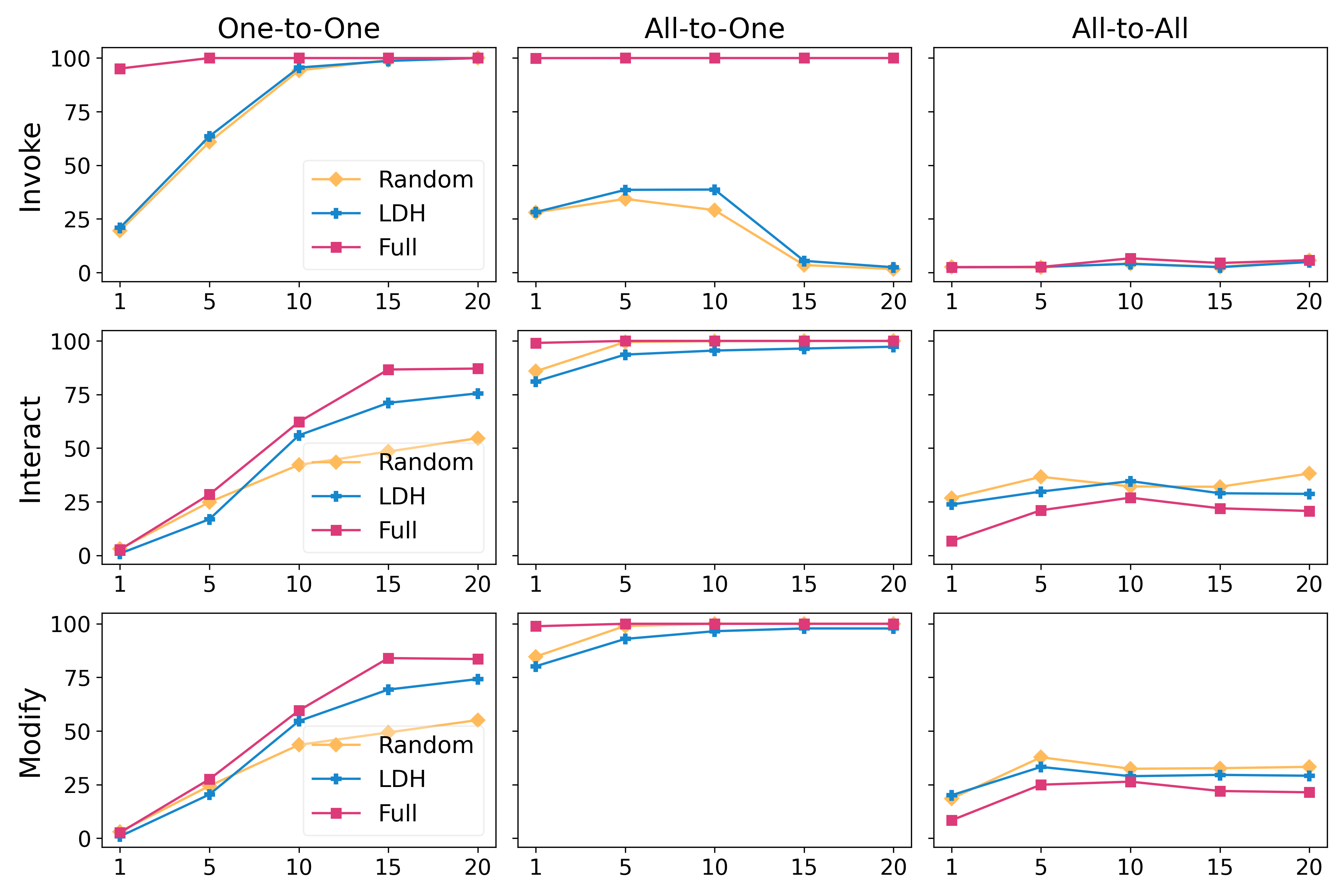}
  \caption{Impact of trigger size in Krait, x-axis denotes the trigger size while y-axis denotes the ASR.}
  \label{tri_size}
  \vspace{-8mm}
\end{figure}

\noindent\textbf{Poisoning rate.} We vary PR as $[2,5,10,20,50]\%$. From Figure \ref{pr}, generally, higher PR, higher ASR. Except for all-to-all attacks, the rest of the attacks with diverse trigger generation methods all accomplish $100\%$ ASRs with less than $5\%$ PR. 
\begin{figure}[ht]
  \centering
  \includegraphics[width=\linewidth]{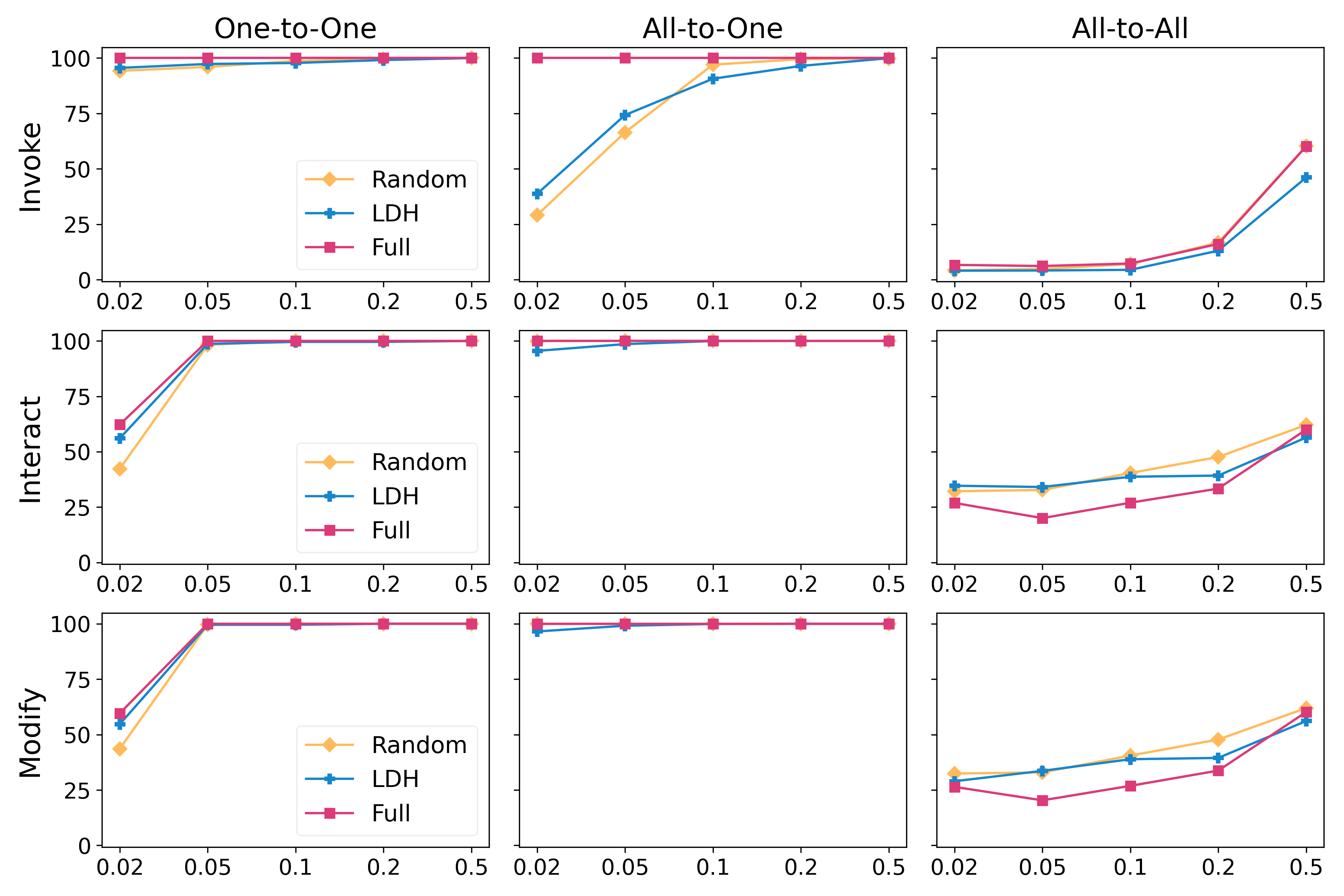}
  \caption{Impact of poisoning rate in Krait, x-axis denotes the PR while y-axis denotes the ASR.}
  \label{pr}
  \vspace{-7mm}
\end{figure}

\noindent\textbf{Training epochs.} We vary training epochs as $[2,4,6,8,10]$. As shown in Figure \ref{train_epoch}, generally, larger epochs yield higher ASRs. In one-to-one attacks with the ``Invoke" method, only training for $2$ epochs with $2$ poisoned nodes yields a $100\%$ ASR, and in all-to-one attacks, within $4$ training epochs, Krait achieves a $100\%$ ASR with $22$ poisoned nodes.
\vspace{-2mm}
\begin{figure}[ht]
  \centering
  \includegraphics[width=\linewidth]{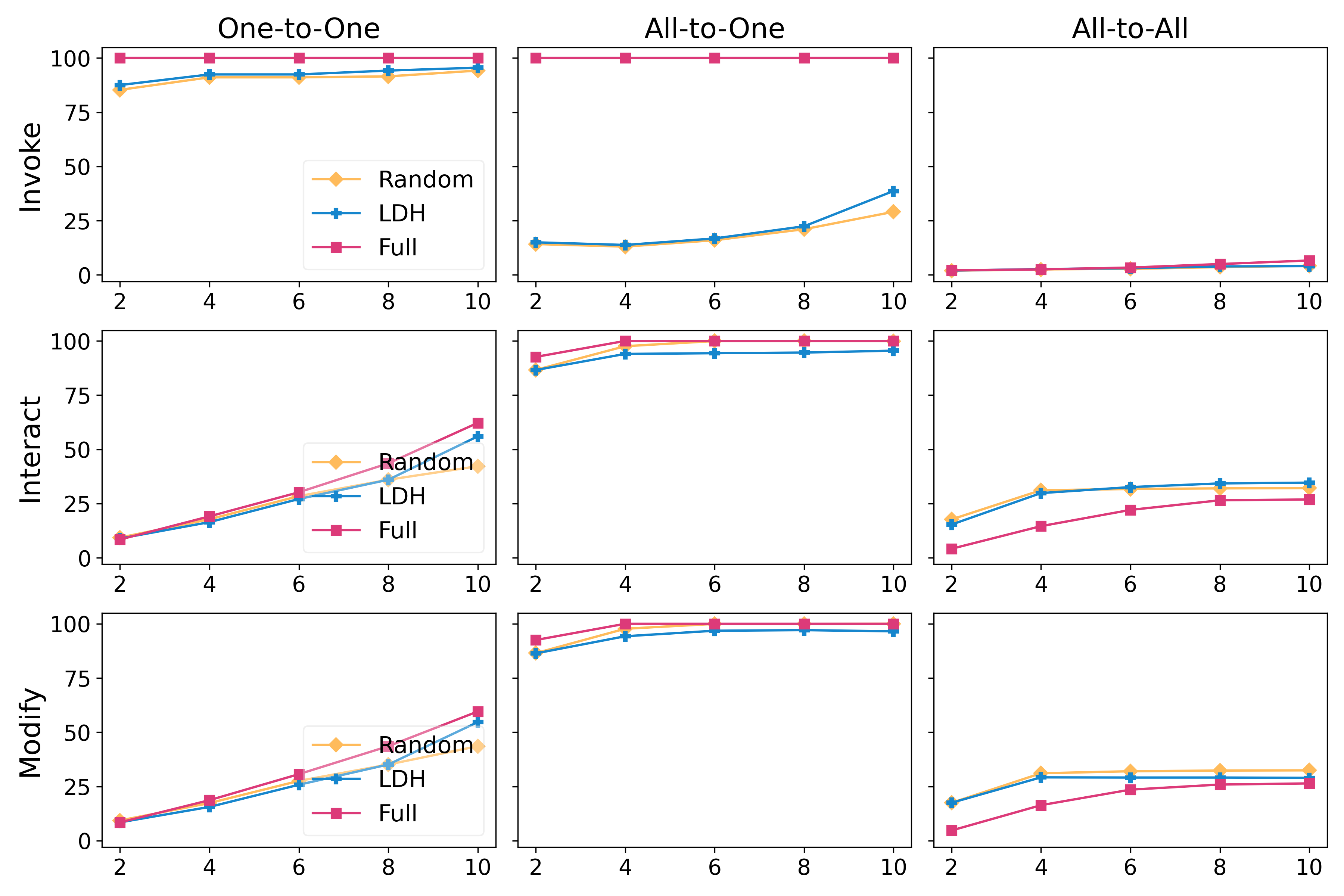}
  \caption{Impact of training epochs in Krait, x-axis denotes the PR while y-axis denotes the ASR.}
  \label{train_epoch}
  \vspace{-9mm}
\end{figure}

\noindent\textbf{Constraint coefficient.}
We vary $\alpha$ as $[0.1, 0.5, 1, 2, 5]$. From Figure \ref{cons_coeff}, the three trigger generation methods perform differently for various attack types. In one-to-one attacks, ``Invoke" surpasses with $100\%$ ASR while ``Interact" and ``Modify" behave almost the same, where their ASRs first increase and then decrease. Similar patterns are observed in all-to-one attacks, but they all maintain $100\%$ ASR with $\alpha$ less than $2$. In all-to-one attacks, the ASRs for ``Invoke" monotonically increase, whereas those for ``Interact'' and ``Modify'' both decline.  
\begin{figure}[ht]
  \centering
  \includegraphics[width=\linewidth]{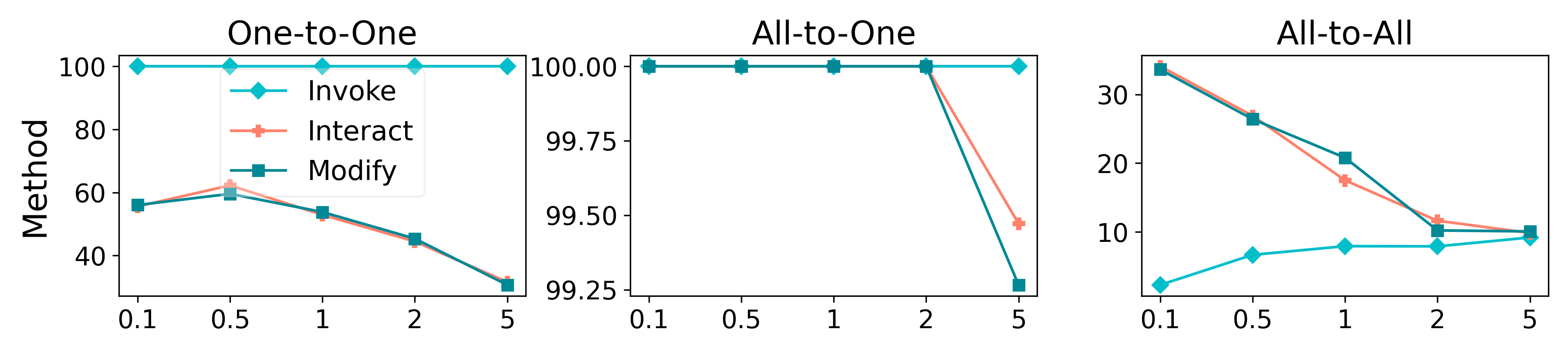}
  \caption{Impact of the coefficient alpha in Krait, x-axis denotes the alpha while y-axis denotes the ASR.}
  \label{cons_coeff}
  \vspace{-9mm}
\end{figure}

\vspace{-2mm}
\subsection{Black-box Setting}
We consider a more practical setting where the system or platform does not support custom training and there is no access to intermediate results. To address this black-box setting, Krait can be degraded into the ``Invoke" trigger generation method without the constraint. We leverage All-in-one with different GNN backbones to mimic the open-sourced models in poisoning warm-ups. We set $\alpha\in$(0.1,0.6). Due to the harsh setting, we set the PR as $10\%$, which only poisons $9$ nodes in one-to-one attacks and $96$ nodes in all-to-one attacks ($0.67\%,7\%$ of the whole training graph, respectively), still much lower than the existing standards. The results are shown in Table \ref{black_Krait}. Under diverse GNN backbones, the black-box Krait still yields comparable performance in terms of attack effectiveness and stealthiness. Krait with the same GT backbone (but with different parameters) achieves superior performance in one-to-one and all-to-one attacks, and Krait with the GCN backbone excels in all-to-all attacks with a $35.95\%$ ASR but it leads to large homophily drops. We further discuss the bad-case analysis in Appendix \ref{bad_black}.

\begin{table*}[!htbp]
\caption{Performance of Krait under the black-box setting. Cite2Cora: CiteSeer to Cora transfer.}
\resizebox{\linewidth}{!}{
\begin{tabular}{c|c|c|cccc|cccc|cccc}
\hline
\textbf{Transfer}                  & \multirow{3}{*}{\textbf{Backbone}}   & \multirow{3}{*}{\textbf{Model}} & \multicolumn{4}{c|}{\textbf{One-to-One}}                                                      & \multicolumn{4}{c|}{\textbf{All-to-One}}                                                      & \multicolumn{4}{c}{\textbf{All-to-All}}                                                      \\ \cline{1-1} \cline{4-15} 
(Evaluation)                   &                           &                        & \multicolumn{2}{c|}{\textbf{Effectiveness}} & \multicolumn{2}{c|}{\textbf{Stealthiness}} & \multicolumn{2}{c|}{\textbf{Effectiveness}} & \multicolumn{2}{c|}{\textbf{Stealthiness}} & \multicolumn{2}{c|}{\textbf{Effectiveness}} & \multicolumn{2}{c}{\textbf{Stealthiness}} \\ \cline{1-1} \cline{4-15} 
(Metrics)                      &                           &                        & ASR       & \multicolumn{1}{c|}{AMC}      & CA                   & AHD         & ASR       & \multicolumn{1}{c|}{AMC}      & CA                   & AHD         & ASR       & \multicolumn{1}{c|}{AMC}      & CA                  & AHD         \\ \hline
\multirow{9}{*}{\textbf{Cite2Cora}} & \multirow{3}{*}{\textbf{GCN}}   & \textbf{Random}                 & 34.22          & 78.65                              & \textbf{86.83}                        & 9.53           & 32.13          & 93.48                              & \textbf{85.07}                        & 9.63            & \textbf{35.95}          & \textbf{95.25}                              & 85.48                        & 19.77            \\
                               &                           & \textbf{LNH}                    & \textbf{43.56}          & \textbf{85.39}                              & 86.81                         & \textbf{9.41}            & 44.88          & 92.50                              & 84.33                        & \textbf{9.36}            & 31.45          & 90.80                              & 86.10                        & \textbf{19.65}            \\
                               &                           & \textbf{Full}                  & 41.33          & 70.16                              & 86.81                         & 14.59            & \textbf{98.41}          & \textbf{99.07}                             & 84.59                        & 11.45            & 30.25          & 91.54                              & \textbf{86.16}                        & 19.96            \\ \cline{2-3}
                               & \multirow{3}{*}{\textbf{GAT}} & \textbf{Random}                 & 4.44          & 35.36                             & \textbf{87.10}                         & -1.09            & 41.00          & 94.46                              & \textbf{84.92}                         & 4.16            & \textbf{13.01}          & \textbf{76.34}                              & 85.60                        & 3.40            \\
                               &                           & \textbf{LNH}                    & 8.89         & 45.90                              & 87.04                         & \textbf{-1.12}            & 51.07          & 92.82                              & 84.21                         & 4.18            & 6.78          & 71.49                              & \textbf{85.83}                       & 3.35            \\
                               &                           & \textbf{Full}                  & \textbf{26.67}          & \textbf{67.60}                              & 86.97                         & 1.79            & \textbf{71.48}          & \textbf{94.47}                              & 84.33                         & \textbf{2.93}            & 6.23          & 70.94                              & 85.71                        & \textbf{2.95}            \\ \cline{2-3}
                               & \multirow{3}{*}{\textbf{GT}}   & \textbf{Random}                 & 98.67               & 96.67                         & \textbf{87.00}                             & 0.34              & 97.03               & 97.35                         & 82.92                            & 4.89             & 7.02               & 78.51                         & \textbf{85.95}                            & 9.72             \\
                                   &                           & \textbf{LNH}                    & 97.78               & 95.04                         & 86.90                            & \textbf{0.30}              & 90.66               & 95.23                         & 83.12                           & 4.91                & 4.43               & 77.81                         & 85.94                           & \textbf{9.58}             \\
                                   &                           & \textbf{Full}                  & \textbf{100.00}               & \textbf{100.00}                         & 86.95                            & 2.73              & \textbf{100.00}               & \textbf{100.00}                         & \textbf{85.71}                             & \textbf{3.08}             & \textbf{8.81}               & \textbf{85.03}                         & 85.66                            & 10.32              \\ \hline
\end{tabular}}
\label{black_Krait}
\vspace{-4mm}
\end{table*}

\vspace{-3mm}
\subsection{Discussions}
\label{discuss_def}
\vspace{-3mm}
\noindent\subsubsection{\textbf{Possible Defense Solutions}}

\begin{table*}[htbp]
\caption{Performance of Krait on GNN-SVD, Noisy-Fea and Noisy-Emb. Cite2Cora: CiteSeer to Cora transfer.}
\resizebox{\linewidth}{!}{
\begin{tabular}{c|c|c|cccc|cccc|cccc}
\toprule
\textbf{Transfer}                      & \multirow{3}{*}{\textbf{Method}}   & \multirow{3}{*}{\textbf{Model}} & \multicolumn{4}{c|}{\textbf{One-to-One}}                                                            & \multicolumn{4}{c|}{\textbf{All-to-One}}                                                           & \multicolumn{4}{c}{\textbf{All-to-All}}                                                            \\ \cline{1-1} \cline{4-15} 
(Evaluation)                       &                           &                        & \multicolumn{2}{c|}{\textbf{Effectiveness}} & \multicolumn{2}{c|}{\textbf{Stealthiness}}       & \multicolumn{2}{c|}{\textbf{Effectiveness}} & \multicolumn{2}{c|}{\textbf{Stealthiness}}      & \multicolumn{2}{c|}{\textbf{Effectiveness}} & \multicolumn{2}{c}{\textbf{Stealthiness}}       \\ \cline{1-1} \cline{4-15} 
(Metrics)                          &                           &                        & ASR            & \multicolumn{1}{c|}{AMC} & CA                        & AHD            & ASR             & \multicolumn{1}{c|}{AMC} & CA                        & AHD           & ASR             & \multicolumn{1}{c|}{AMC} & CA                       & AHD           \\ \hline
\multirow{9}{*}{\textbf{Cite2Cora}}     
& \multirow{3}{*}{\textbf{Invoke}}   
                                  & \textbf{GNN-SVD}                 
                                  & 100.00               & 100.00                         & \textbf{86.77}                            & \textbf{7.71}              
                                  & 100.00               & 100.00                         & 85.83                           & \textbf{6.02}             
                                  & 6.90               & 82.50                         & \textbf{86.13}              & \textbf{15.72}             \\
                                  & 
                                   & \textbf{Noisy-Fea}                    
                                   & 100.00               & 100.00                         & 86.72                             & 3.23              
                                   & 100.00               & 100.00                         & \textbf{85.95}                            & 3.54             
                                   & \textbf{6.11}               & \textbf{80.71}                         & 86.07            & 10.54   
                                   \\
                                   &                           
                                   & \textbf{Noisy-Emb}                    
                                   & 100.00               & 100.00                         & 86.75                            & 3.02              
                                   & 100.00               & 100.00                         & 85.83                           & 3.27             
                                   & 6.75               & 81.74                         & 86.07                         & 10.44    \\    
                                    \cline{2-3}
& \multirow{3}{*}{\textbf{Interact}} 
                                   & \textbf{GNN-SVD}                 
                                   & \textbf{61.78} & 91.90           & 86.69  & \textbf{7.89} 
                                   & 100.00 & 100.00            & 85.89  & \textbf{7.91} 
                                   & \textbf{31.01}  & \textbf{89.74}            & 86.16  & \textbf{11.29} \\
                                   & 
                                   & \textbf{Noisy-Fea}                    
                                   & 64.44               & \textbf{89.72}                         & 86.65                             & 1.60              
                                   & 100.00               & 100.00                         & \textbf{86.13}                             & 4.32             
                                   & 31.45               & 90.71                         & 86.16                             & 6.07                  
                                   \\
                                   &                           
                                   & \textbf{Noisy-Emb}                    
                                   & 63.56 & 90.65           & \textbf{86.71}  & 1.29 
                                   & 100.00  & 100.00            & 85.86  & 4.05 
                                   & 31.89               & 88.47                         & \textbf{86.22}                             & 5.96        \\    
                                    \cline{2-3}
& \multirow{3}{*}{\textbf{Modify}}   
                                   & \textbf{GNN-SVD}                 
                                   & \textbf{60.89}               & 90.20                         & \textbf{86.69}                         & \textbf{5.63}              
                                   & 100.00               & 100.00                         & 85.92                             & \textbf{6.61}             
                                   & \textbf{31.60}               & \textbf{89.50}                         & \textbf{86.27}  & \textbf{9.83}             \\
                                   & 
                                   & \textbf{Noisy-Fea}                    
                                   & 64.00               & \textbf{88.27}                         & 86.62                           & 1.61              
                                   & 100.00               & 100.00                         & \textbf{86.01}                       & 4.35             
                                   & 31.78               & 88.99                         & 86.24              
                                   & 6.11                    \\
                                   &                           
                                   & \textbf{Noisy-Emb}                    
                                   & 61.78               & 90.53                         & \textbf{86.69}                     & 1.29              
                                   & 100.00               & 100.00                         & 85.92                         & 4.06             
                                   & 31.81               & 88.61                         & 86.22                       & 5.82        \\    
                                    \hline
\end{tabular}
}
\label{krait_defense}
\vspace{-6mm}
\end{table*}

Given that defense mechanisms against graph backdoor attacks are still in their infancy, we resort to defenses from other graph adversarial attacks and injection-based backdoor detections in different domains to analyze and assess their effectiveness against Krait.

\noindent\textbf{Homophily and node-similarity defenders.}
Chen et al. \cite{unnoticeability_node_centric_homo} propose homophily defenders against graph injection attacks and demonstrate that GNNGuard \cite{gnnguard}, an acclaimed defense mechanism \cite{contrastive_backdoor}, acts as a homophily defender by pruning suspicious edges if $sim(X_v, X_u)\leq s_{\mathcal{G}}$, where $s_{\mathcal{G}}$ denotes the minimum similarity for any pair of nodes connected in $\mathcal{G}$. Nonetheless, GNNGuard becomes ineffective against Krait for the following four reasons: 1) As indicated in Section \ref{sec:prelim_exp}, the consistent homophily patterns before and after graph prompt backdoor injections make it more intractable for defenders to identify poisoned nodes; 2) GNNGuard is primarily designed for node classification and cannot be trivially applied to graph prompt backdoor attacks since the latter reformulate node-level tasks to graph-level tasks, complicating the detection of small-scale trigger injection in each ego-network. Meanwhile, as shown in Table \ref{Krait_perf}-\ref{black_Krait}, homophily variations between clean samples and poisoned nodes are typically slight; 3) The structure of prompt tokens, which are mainly similarity functions of token features, can be adaptively tuned when crafting poisoned prompts, making them more difficult for homophily defenders to defeat. This challenge likewise extends to node similarity-based detectors mentioned in Zheng et al. \cite{motif_backdoor}; 4) Ennadir et al. \cite{NoisyGNN} empirically validate that homophily defenders, i.e., GNNGuard, are $11.78-303.43$ times slower than other defense mechanisms. GNN-Jaccard \cite{gnn_jaccard}, which utilizes Jaccard similarity (only count $0$ or $1$) to remove suspicious edges if their linked nodes' similarity scores are close to $0$, will lose efficacy since the myriad of graph prompt tuning models \cite{all_in_one} employ dimensionality reduction techniques, especially in large-scale graphs with nearly $10k$ node features \cite{physics_computers}. 

\noindent\textbf{Model inspection.} Xi et al. \cite{graph_backdoor} extend NeuralCleanse \cite{neural_cleanse} to graph learning, which computes graph edit distance between triggers and to-be-replaced subgraphs. Krait, however, eliminates the necessity for subgraph replacement, thus restricting backdoor detection. Even with the allowance, model inspection will attempt to search a specific trigger in every subgraph for $|\mathcal{V}|$ times, which is impractical for large-scale graphs.

\noindent\textbf{Randomized smoothing.} Randomized smoothing\cite{random_backdoor} is a potential defense mechanism, which samples numerous subgraphs of each ego-network for a majority voting. However, defending against graph prompt backdoors is impractical because the sampling operation itself has a complexity of $\mathcal{O}(|\mathcal{V}|\cdot \mathcal{C}_{s})$, where $\mathcal{C}_{s}$ is the original complexity of sampling methods. Not to mention the more time-intensive training for all sampled subgraphs. The total computational costs are prohibitive.

\vspace{-2mm}
\noindent\textbf{Noise filtering and injection.} Another line of defenders manipulates noises to fight against adversarial perturbations. Entezari et al. \cite{GNNSVD} demonstrate that graph adversarial attacks inject noises into higher ranks in the singular value spectrum. Therefore, they design GNN-SVD, which employs a low-rank approximation of the adjacency matrix to filter out high-rank noises and reset the remaining values to $1$, thus completing the pruning of noisy edges. Inversely, NoisyGNN \cite{NoisyGNN} injects noises from a pre-defined distribution into the node feature/structural space to defend against feature/structural-based adversarial attacks. To align with Krait, we implement GNN-SVD and NoisyGNN for each subgraph. We follow GNN-SVD's recommendation to set rank as $10$. As for NoisyGNN, we sample noises from Gaussian distribution $\mathcal{N}(0, \mathcal{I})$ with the default scaling parameter as $0.1$. We first inject noises into the feature spaces of the prompted subgraphs (marked as Noisy-Fea) as node features play vital roles in graph prompt tuning. Next, since NoisyGNN is designed for node classification and cannot be trivially adapted to graph prompt tuning which involves node-subgraph transformations, we consider a new variant of NoisyGNN (Noisy-Emb) to inject noises into the graph embeddings of the prompted subgraphs. The defense performance is shown in Table \ref{krait_defense}, indicating these two methods cannot effectively defend against Krait. Interestingly, in some specific cases, such as in the one-to-one attacks with ``Modify" as the trigger generation method, GNN-SVD and the two variants of NoisyGNN actually enhance the attack performance of Krait. We speculate that 1) GNN-SVD may filter out noises that hinder the creation of a solid mapping between triggers and the target label, thus improving attack effectiveness; 2) By adding noises into feature and hidden spaces of poisoned graphs, NoisyGNN reinforces the robustness of Krait, leading to more stable and accurate performance. We leave the exploration of the mechanisms behind this phenomenon for future work.
\subsubsection{\textbf{Suggestions for Graph Prompt Protection}}

Wang et al. \cite{pre_conf_defense} indicate that backdoor attacks often yield statistically higher confidence scores of poisoned samples to achieve high ASRs with relatively low PRs. Although it \cite{pre_conf_defense} focuses on backdoor attacks in CV, we observe similar patterns of Krait in Table \ref{Krait_perf}-\ref{black_Krait}. Besides, since Krait utilizes graph prompt tuning to craft graph prompts as triggers, the ego-networks of poisoned candidates typically exhibit larger average degrees compared to the clean samples. These evidences and observations inspire us to monitor suspicious nodes with unusually higher confidence scores and larger average degrees. However, previous studies in fair graph learning suggest that nodes with larger degrees enjoy more benefits during message-passing processes, resulting in higher confidence scores \cite{degree_bias,degree_fairness}. Therefore, filtering such suspicious nodes based solely on the above criteria could impair the performance of graph prompt tuning. The same conclusion can be inferred from Figure \ref{conf_score_dist}, where the overwhelming majority of clean samples, as well as the poisoned samples, have confidence scores closer to $1$. Nevertheless, it is still valuable to monitor nodes with high confidence scores. 

\vspace{-2mm}
\begin{figure}[htbp]
  \centering
  \includegraphics[width=\linewidth]{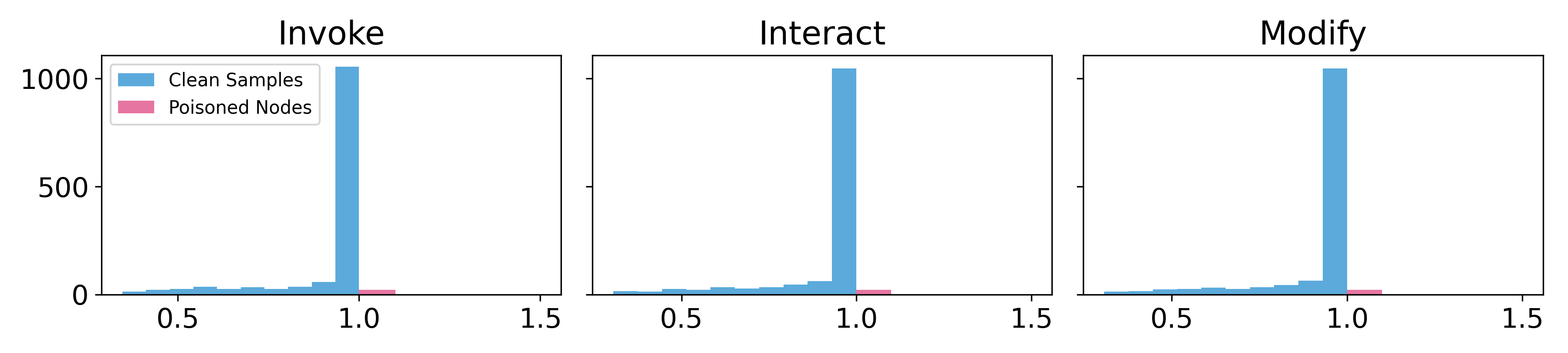}
  \vspace{-6mm}
  \caption{Confidence score distributions of clean and poisoned samples under three trigger generation methods.}
  \label{conf_score_dist}
  \vspace{-4mm}
\end{figure}

Another possible direction is to project suspiciously backdoored embeddings into lower dimensions to identify potential anomalies. To validate whether anomalies exist, we visualize the node embeddings of one-to-one and all-to-one attacks in Figure \ref{prompt_backdoor_viz}-\ref{prompt_backdoor_viz_modify_appendix} in Appendix \ref{Backdoored_emb_viz_appendix}. We observe that the poisoned samples tend to cluster at the edges of class boundaries. Therefore, tracing nodes with high confidence scores and proximity to decision boundaries could be an effective strategy. 
Moreover, to prevent graph prompt backdoors that utilize ``Interact" and ``Modify" trigger generation methods, service providers or model/system owners should prohibit the generation of multiple graph prompts for arbitrary node sets. They should also implement rigorous scrutiny and auditing practices to ensure that these prompts and their sources have not been maliciously altered. Moreover, we recommend prioritizing system-automated graph prompts over potentially suspicious hand-crafted prompts, especially when anomalies are detected.

\subsubsection{\textbf{Limitations and Future Work}}
Although we showcase the effectiveness and stealthiness of Krait, several limitations still exist. First, we anticipate testing Krait in real-world APIs or platforms. Second, we utilize label flipping for poisoned nodes \cite{graph_backdoor, unnoticeable_graph_backdoor, explain_backdoor}, developing a clean-label Krait can be a promising direction. Third, we only evaluate Krait in homophilic graphs, generating heterogeneous graph prompts as triggers can enhance its impacts. Fourth, Krait can be easily extended to multiple graph prompts, which better reflects real-world scenarios. Different attackers can handcraft or hijack different graph prompts at various times, leading to much more severe social impacts. Therefore, this problem should be further highlighted. Finally, it is critical to develop a more reliable defense mechanism against graph prompt backdoors.
\vspace{-2mm}
\section{Related Work}
\subsection{Graph Prompt Tuning} 

GPPT \cite{GPPT} proposes a graph prompting function to transform each node into a token pair, thereby reformulating downstream node-level tasks in line with the pretext. GraphPrompt \cite{graph_prompts} utilizes subgraph similarity to unify pre-training and downstream tasks into a common task template and adopts learnable prompts to encode the most relevant graph knowledge in a task-specific manner. PRODIGY \cite{prodigy} introduces in-context learning on graphs, employing hand-crafted graph prompts without tunable parameters to represent node, edge, and graph-level tasks. GPF and GPF-plus \cite{graph_prompt_theory} use task-specific tuning to learn a universal feature vector as the prompt, which can be generalized to various pre-training and downstream task adaptations. All-in-one \cite{all_in_one} aligns all-level tasks into the graph-level and treats additional subgraphs as graph prompts, where prompt features can be tuned through meta-learning, and the inner structures and inserting patterns are functions of the prompt features and node features, respectively.

\vspace{-2mm}
\subsection{Graph Backdoor Attacks}
Zhang et al. \cite{random_backdoor} construct random subgraphs as triggers and then attach these triggers to random nodes. Xi et al. \cite{graph_backdoor} utilize bi-level optimization to generate triggers for specific downstream tasks while implanting these triggers into random nodes. Xu et al. \cite{explain_backdoor} employ explanation-based methods to select the least important nodes as poisoned candidates, randomly altering their features to create feature triggers. Zheng et al. \cite{motif_backdoor} combine motif distributions with subgraph importance to design graph backdoors, demonstrating that using low-importance subgraphs can achieve better attack performance. Dai et al. \cite{unnoticeable_graph_backdoor} first identify poisoned samples through node clustering, then leverage bi-level optimization to generate triggers that can survive under pre-processing defense mechanisms. Zhao et al. \cite{spectral_backdoor} develop spectral graph backdoorsby choosing significant feature and frequency bands for injecting trigger signals. Zhang et al. \cite{contrastive_backdoor} propose the first graph backdoor attack against graph contrastive learning (GCL), which uses discrete optimization to attack possible stages of the GCL pipeline. Xu et al. \cite{multi_gnn_backdoor} consider one-to-N and N-to-one attacks, which flip labels of randomly sampled nodes and modify their features to specific values. However, N is set to a small number, such as $2$ or $4$, which cannot cover all node labels needed to launch more advanced one-to-all or all-to-all attacks.


\vspace{-2mm}
\section{Conclusion}

\label{sec:conclusion}

In this study, we pioneer the investigation into the inherent vulnerability of graph prompt tuning. We introduce Krait, a novel graph prompt backdoor that utilizes label non-uniformity homophily to effectively select poisoned nodes during the pre-training phase. Additionally, we design three trigger generation methods to treat graph prompts as triggers in distinct scenarios and incorporate a novel centroid similarity-based loss function to enhance the effectiveness and stealthiness of Krait. Our experimental results validate the practicality of Krait across various combinations of attack types, trigger generation methods, and GNN backbones. Moreover, Krait can be successfully extended to the more challenging black-box setting. Finally, we analyze and assess potential countermeasures on Krait, and further address the necessity to develop a reliable defense mechanism against graph prompt backdoor attacks.







\section*{Acknowledgment}
This research was supported in part by the University of Pittsburgh Center for Research Computing through the resources provided. The first author acknowledges the support from the SCI fellowship.

\bibliographystyle{IEEEtranS}
\bibliography{Reference}

\begin{thebibliography}{10}
\providecommand{\url}[1]{#1}
\csname url@samestyle\endcsname
\providecommand{\newblock}{\relax}
\providecommand{\bibinfo}[2]{#2}
\providecommand{\BIBentrySTDinterwordspacing}{\spaceskip=0pt\relax}
\providecommand{\BIBentryALTinterwordstretchfactor}{4}
\providecommand{\BIBentryALTinterwordspacing}{\spaceskip=\fontdimen2\font plus
\BIBentryALTinterwordstretchfactor\fontdimen3\font minus \fontdimen4\font\relax}
\providecommand{\BIBforeignlanguage}[2]{{%
\expandafter\ifx\csname l@#1\endcsname\relax
\typeout{** WARNING: IEEEtranS.bst: No hyphenation pattern has been}%
\typeout{** loaded for the language `#1'. Using the pattern for}%
\typeout{** the default language instead.}%
\else
\language=\csname l@#1\endcsname
\fi
#2}}
\providecommand{\BIBdecl}{\relax}
\BIBdecl

\bibitem{quick_select}
\BIBentryALTinterwordspacing
A.~Alexandrescu, ``Fast deterministic selection,'' 2016. [Online]. Available: \url{https://arxiv.org/abs/1606.00484}
\BIBentrySTDinterwordspacing

\bibitem{unnoticeability_node_centric_homo}
\BIBentryALTinterwordspacing
Y.~Chen, H.~Yang, Y.~Zhang, M.~KAILI, T.~Liu, B.~Han, and J.~Cheng, ``Understanding and improving graph injection attack by promoting unnoticeability,'' in \emph{International Conference on Learning Representations}, 2022. [Online]. Available: \url{https://openreview.net/forum?id=wkMG8cdvh7-}
\BIBentrySTDinterwordspacing

\bibitem{unnoticeable_graph_backdoor}
\BIBentryALTinterwordspacing
E.~Dai, M.~Lin, X.~Zhang, and S.~Wang, ``Unnoticeable backdoor attacks on graph neural networks,'' in \emph{Proceedings of the ACM Web Conference 2023}, ser. WWW '23.\hskip 1em plus 0.5em minus 0.4em\relax New York, NY, USA: Association for Computing Machinery, 2023, p. 2263–2273. [Online]. Available: \url{https://doi.org/10.1145/3543507.3583392}
\BIBentrySTDinterwordspacing

\bibitem{all_to_one_all}
K.~D. Doan, Y.~Lao, and P.~Li, ``Marksman backdoor: backdoor attacks with arbitrary target class,'' in \emph{Proceedings of the 36th International Conference on Neural Information Processing Systems}, ser. NIPS '22.\hskip 1em plus 0.5em minus 0.4em\relax Red Hook, NY, USA: Curran Associates Inc., 2024.

\bibitem{NoisyGNN}
\BIBentryALTinterwordspacing
S.~Ennadir, Y.~Abbahaddou, J.~F. Lutzeyer, M.~Vazirgiannis, and H.~Boström, ``A simple and yet fairly effective defense for graph neural networks,'' \emph{Proceedings of the AAAI Conference on Artificial Intelligence}, vol.~38, no.~19, pp. 21\,063--21\,071, Mar. 2024. [Online]. Available: \url{https://ojs.aaai.org/index.php/AAAI/article/view/30098}
\BIBentrySTDinterwordspacing

\bibitem{GNNSVD}
\BIBentryALTinterwordspacing
N.~Entezari, S.~A. Al-Sayouri, A.~Darvishzadeh, and E.~E. Papalexakis, ``All you need is low (rank): Defending against adversarial attacks on graphs,'' in \emph{Proceedings of the 13th International Conference on Web Search and Data Mining}, ser. WSDM '20.\hskip 1em plus 0.5em minus 0.4em\relax New York, NY, USA: Association for Computing Machinery, 2020, p. 169–177. [Online]. Available: \url{https://doi.org/10.1145/3336191.3371789}
\BIBentrySTDinterwordspacing

\bibitem{recomm_app}
\BIBentryALTinterwordspacing
W.~Fan, Y.~Ma, Q.~Li, Y.~He, E.~Zhao, J.~Tang, and D.~Yin, ``Graph neural networks for social recommendation,'' in \emph{The World Wide Web Conference}, ser. WWW '19.\hskip 1em plus 0.5em minus 0.4em\relax New York, NY, USA: Association for Computing Machinery, 2019, p. 417–426. [Online]. Available: \url{https://doi.org/10.1145/3308558.3313488}
\BIBentrySTDinterwordspacing

\bibitem{graph_prompt_theory}
\BIBentryALTinterwordspacing
T.~Fang, Y.~Zhang, Y.~YANG, C.~Wang, and L.~Chen, ``Universal prompt tuning for graph neural networks,'' in \emph{Advances in Neural Information Processing Systems}, A.~Oh, T.~Neumann, A.~Globerson, K.~Saenko, M.~Hardt, and S.~Levine, Eds., vol.~36.\hskip 1em plus 0.5em minus 0.4em\relax Curran Associates, Inc., 2023, pp. 52\,464--52\,489. [Online]. Available: \url{https://proceedings.neurips.cc/paper_files/paper/2023/file/a4a1ee071ce0fe63b83bce507c9dc4d7-Paper-Conference.pdf}
\BIBentrySTDinterwordspacing

\bibitem{b3_black_box_backdoor}
\BIBentryALTinterwordspacing
X.~Gong, Y.~Chen, W.~Yang, H.~Huang, and Q.~Wang, ``B3: Backdoor attacks against black-box machine learning models,'' \emph{ACM Trans. Priv. Secur.}, vol.~26, no.~4, aug 2023. [Online]. Available: \url{https://doi.org/10.1145/3605212}
\BIBentrySTDinterwordspacing

\bibitem{first_work}
T.~Gu, K.~Liu, B.~Dolan-Gavitt, and S.~Garg, ``Badnets: Evaluating backdooring attacks on deep neural networks,'' \emph{IEEE Access}, vol.~7, pp. 47\,230--47\,244, 2019.

\bibitem{prodigy}
\BIBentryALTinterwordspacing
Q.~Huang, H.~Ren, P.~Chen, G.~Kr{\v{z}}manc, D.~Zeng, P.~Liang, and J.~Leskovec, ``{PRODIGY}: Enabling in-context learning over graphs,'' in \emph{Thirty-seventh Conference on Neural Information Processing Systems}, 2023. [Online]. Available: \url{https://openreview.net/forum?id=pLwYhNNnoR}
\BIBentrySTDinterwordspacing

\bibitem{label_non_uni}
F.~Ji, S.~H. Lee, H.~Meng, K.~Zhao, J.~Yang, and W.~P. Tay, ``Leveraging label non-uniformity for node classification in graph neural networks,'' in \emph{Proceedings of the 40th International Conference on Machine Learning}, ser. ICML'23.\hskip 1em plus 0.5em minus 0.4em\relax JMLR.org, 2023.

\bibitem{GCN}
\BIBentryALTinterwordspacing
T.~N. Kipf and M.~Welling, ``Semi-supervised classification with graph convolutional networks,'' in \emph{International Conference on Learning Representations}, 2017. [Online]. Available: \url{https://openreview.net/forum?id=SJU4ayYgl}
\BIBentrySTDinterwordspacing

\bibitem{multi_prompt_recommendation}
\BIBentryALTinterwordspacing
L.~Li, Y.~Zhang, and L.~Chen, ``Prompt distillation for efficient llm-based recommendation,'' in \emph{Proceedings of the 32nd ACM International Conference on Information and Knowledge Management}, ser. CIKM '23.\hskip 1em plus 0.5em minus 0.4em\relax New York, NY, USA: Association for Computing Machinery, 2023, p. 1348–1357. [Online]. Available: \url{https://doi.org/10.1145/3583780.3615017}
\BIBentrySTDinterwordspacing

\bibitem{bio_health}
\BIBentryALTinterwordspacing
M.~M. Li, K.~Huang, and M.~Zitnik, ``Graph representation learning in biomedicine and healthcare,'' \emph{Nature Biomedical Engineering}, vol.~6, pp. 1353 -- 1369, 2022. [Online]. Available: \url{https://api.semanticscholar.org/CorpusID:253245343}
\BIBentrySTDinterwordspacing

\bibitem{degree_fairness}
\BIBentryALTinterwordspacing
Z.~Liu, T.-K. Nguyen, and Y.~Fang, ``On generalized degree fairness in graph neural networks,'' in \emph{Proceedings of the Thirty-Seventh AAAI Conference on Artificial Intelligence and Thirty-Fifth Conference on Innovative Applications of Artificial Intelligence and Thirteenth Symposium on Educational Advances in Artificial Intelligence}, ser. AAAI'23/IAAI'23/EAAI'23.\hskip 1em plus 0.5em minus 0.4em\relax AAAI Press, 2023. [Online]. Available: \url{https://doi.org/10.1609/aaai.v37i4.25574}
\BIBentrySTDinterwordspacing

\bibitem{graph_prompts}
\BIBentryALTinterwordspacing
Z.~Liu, X.~Yu, Y.~Fang, and X.~Zhang, ``Graphprompt: Unifying pre-training and downstream tasks for graph neural networks,'' in \emph{Proceedings of the ACM Web Conference 2023}, ser. WWW '23.\hskip 1em plus 0.5em minus 0.4em\relax New York, NY, USA: Association for Computing Machinery, 2023, p. 417–428. [Online]. Available: \url{https://doi.org/10.1145/3543507.3583386}
\BIBentrySTDinterwordspacing

\bibitem{homo_class}
\BIBentryALTinterwordspacing
S.~Luan, C.~Hua, M.~Xu, Q.~Lu, J.~Zhu, X.-W. Chang, J.~Fu, J.~Leskovec, and D.~Precup, ``When do graph neural networks help with node classification? investigating the homophily principle on node distinguishability,'' in \emph{Thirty-seventh Conference on Neural Information Processing Systems}, 2023. [Online]. Available: \url{https://openreview.net/forum?id=kJmYu3Ti2z}
\BIBentrySTDinterwordspacing

\bibitem{homo_class1}
\BIBentryALTinterwordspacing
Y.~Ma, X.~Liu, N.~Shah, and J.~Tang, ``Is homophily a necessity for graph neural networks?'' in \emph{International Conference on Learning Representations}, 2022. [Online]. Available: \url{https://openreview.net/forum?id=ucASPPD9GKN}
\BIBentrySTDinterwordspacing

\bibitem{healthcare_graph_prompt}
L.~Peng, S.~Cai, Z.~Wu, H.~Shang, X.~Zhu, and X.~Li, ``Mmgpl: Multimodal medical data analysis with graph prompt learning,'' 2023.

\bibitem{physics_computers}
O.~Shchur, M.~Mumme, A.~Bojchevski, and S.~Günnemann, ``Pitfalls of graph neural network evaluation,'' 2019.

\bibitem{graph_transformer}
\BIBentryALTinterwordspacing
Y.~Shi, Z.~Huang, S.~Feng, H.~Zhong, W.~Wang, and Y.~Sun, ``Masked label prediction: Unified message passing model for semi-supervised classification,'' in \emph{Proceedings of the Thirtieth International Joint Conference on Artificial Intelligence, {IJCAI-21}}, Z.-H. Zhou, Ed.\hskip 1em plus 0.5em minus 0.4em\relax International Joint Conferences on Artificial Intelligence Organization, 8 2021, pp. 1548--1554, main Track. [Online]. Available: \url{https://doi.org/10.24963/ijcai.2021/214}
\BIBentrySTDinterwordspacing

\bibitem{GPPT}
\BIBentryALTinterwordspacing
M.~Sun, K.~Zhou, X.~He, Y.~Wang, and X.~Wang, ``Gppt: Graph pre-training and prompt tuning to generalize graph neural networks,'' in \emph{Proceedings of the 28th ACM SIGKDD Conference on Knowledge Discovery and Data Mining}, ser. KDD '22.\hskip 1em plus 0.5em minus 0.4em\relax New York, NY, USA: Association for Computing Machinery, 2022, p. 1717–1727. [Online]. Available: \url{https://doi.org/10.1145/3534678.3539249}
\BIBentrySTDinterwordspacing

\bibitem{all_in_one}
\BIBentryALTinterwordspacing
X.~Sun, H.~Cheng, J.~Li, B.~Liu, and J.~Guan, ``All in one: Multi-task prompting for graph neural networks,'' in \emph{Proceedings of the 29th ACM SIGKDD Conference on Knowledge Discovery and Data Mining}, ser. KDD '23.\hskip 1em plus 0.5em minus 0.4em\relax New York, NY, USA: Association for Computing Machinery, 2023, p. 2120–2131. [Online]. Available: \url{https://doi.org/10.1145/3580305.3599256}
\BIBentrySTDinterwordspacing

\bibitem{graph_prompt_survey}
X.~Sun, J.~Zhang, X.~Wu, H.~Cheng, Y.~Xiong, and J.~Li, ``Graph prompt learning: A comprehensive survey and beyond,'' 2023.

\bibitem{VNT}
\BIBentryALTinterwordspacing
Z.~Tan, R.~Guo, K.~Ding, and H.~Liu, ``Virtual node tuning for few-shot node classification,'' in \emph{Proceedings of the 29th ACM SIGKDD Conference on Knowledge Discovery and Data Mining}, ser. KDD '23.\hskip 1em plus 0.5em minus 0.4em\relax New York, NY, USA: Association for Computing Machinery, 2023, p. 2177–2188. [Online]. Available: \url{https://doi.org/10.1145/3580305.3599541}
\BIBentrySTDinterwordspacing

\bibitem{anomaly_detect}
J.~Tang, J.~Li, Z.~Gao, and J.~Li, ``Rethinking graph neural networks for anomaly detection,'' in \emph{International Conference on Machine Learning}, 2022.

\bibitem{degree_bias}
\BIBentryALTinterwordspacing
X.~Tang, H.~Yao, Y.~Sun, Y.~Wang, J.~Tang, C.~Aggarwal, P.~Mitra, and S.~Wang, ``Investigating and mitigating degree-related biases in graph convoltuional networks,'' in \emph{Proceedings of the 29th ACM International Conference on Information \& Knowledge Management}, ser. CIKM '20.\hskip 1em plus 0.5em minus 0.4em\relax New York, NY, USA: Association for Computing Machinery, 2020, p. 1435–1444. [Online]. Available: \url{https://doi.org/10.1145/3340531.3411872}
\BIBentrySTDinterwordspacing

\bibitem{GAT}
\BIBentryALTinterwordspacing
P.~Veličković, G.~Cucurull, A.~Casanova, A.~Romero, P.~Liò, and Y.~Bengio, ``Graph attention networks,'' in \emph{International Conference on Learning Representations}, 2018. [Online]. Available: \url{https://openreview.net/forum?id=rJXMpikCZ}
\BIBentrySTDinterwordspacing

\bibitem{neural_cleanse}
B.~Wang, Y.~Yao, S.~Shan, H.~Li, B.~Viswanath, H.~Zheng, and B.~Y. Zhao, ``Neural cleanse: Identifying and mitigating backdoor attacks in neural networks,'' in \emph{2019 IEEE Symposium on Security and Privacy (SP)}, 2019, pp. 707--723.

\bibitem{one_to_many_backdoor}
\BIBentryALTinterwordspacing
K.~Wang, H.~Deng, Y.~Xu, Z.~Liu, and Y.~Fang, ``Multi-target label backdoor attacks on graph neural networks,'' \emph{Pattern Recognition}, vol. 152, p. 110449, 2024. [Online]. Available: \url{https://www.sciencedirect.com/science/article/pii/S0031320324002000}
\BIBentrySTDinterwordspacing

\bibitem{pre_conf_defense}
\BIBentryALTinterwordspacing
T.~Wang, Y.~Yao, F.~Xu, M.~Xu, S.~An, and T.~Wang, ``Inspecting prediction confidence for detecting black-box backdoor attacks,'' \emph{Proceedings of the AAAI Conference on Artificial Intelligence}, vol.~38, no.~1, pp. 274--282, Mar. 2024. [Online]. Available: \url{https://ojs.aaai.org/index.php/AAAI/article/view/27780}
\BIBentrySTDinterwordspacing

\bibitem{gnn_jaccard}
\BIBentryALTinterwordspacing
H.~Wu, C.~Wang, Y.~Tyshetskiy, A.~Docherty, K.~Lu, and L.~Zhu, ``Adversarial examples for graph data: Deep insights into attack and defense,'' in \emph{Proceedings of the Twenty-Eighth International Joint Conference on Artificial Intelligence, {IJCAI-19}}.\hskip 1em plus 0.5em minus 0.4em\relax International Joint Conferences on Artificial Intelligence Organization, 7 2019, pp. 4816--4823. [Online]. Available: \url{https://doi.org/10.24963/ijcai.2019/669}
\BIBentrySTDinterwordspacing

\bibitem{graph_backdoor}
Z.~Xi, R.~Pang, S.~Ji, and T.~Wang, ``Graph backdoor,'' in \emph{30th $\{$USENIX$\}$ Security Symposium ($\{$USENIX$\}$ Security 21)}, 2021.

\bibitem{multi_gnn_backdoor}
\BIBentryALTinterwordspacing
J.~Xu and S.~Picek, ``Poster: Multi-target \& multi-trigger backdoor attacks on graph neural networks,'' in \emph{Proceedings of the 2023 ACM SIGSAC Conference on Computer and Communications Security}, ser. CCS '23.\hskip 1em plus 0.5em minus 0.4em\relax New York, NY, USA: Association for Computing Machinery, 2023, p. 3570–3572. [Online]. Available: \url{https://doi.org/10.1145/3576915.3624387}
\BIBentrySTDinterwordspacing

\bibitem{explain_backdoor}
\BIBentryALTinterwordspacing
J.~Xu, M.~J. Xue, and S.~Picek, ``Explainability-based backdoor attacks against graph neural networks,'' in \emph{Proceedings of the 3rd ACM Workshop on Wireless Security and Machine Learning}, ser. WiseML '21.\hskip 1em plus 0.5em minus 0.4em\relax New York, NY, USA: Association for Computing Machinery, 2021, p. 31–36. [Online]. Available: \url{https://doi.org/10.1145/3468218.3469046}
\BIBentrySTDinterwordspacing

\bibitem{pre-train_backdoor}
\BIBentryALTinterwordspacing
S.~Yang, B.~G. Doan, P.~Montague, O.~De~Vel, T.~Abraham, S.~Camtepe, D.~C. Ranasinghe, and S.~S. Kanhere, ``Transferable graph backdoor attack,'' in \emph{Proceedings of the 25th International Symposium on Research in Attacks, Intrusions and Defenses}, ser. RAID '22.\hskip 1em plus 0.5em minus 0.4em\relax New York, NY, USA: Association for Computing Machinery, 2022, p. 321–332. [Online]. Available: \url{https://doi.org/10.1145/3545948.3545976}
\BIBentrySTDinterwordspacing

\bibitem{cora_citeseer}
Z.~Yang, W.~W. Cohen, and R.~Salakhutdinov, ``Revisiting semi-supervised learning with graph embeddings,'' in \emph{Proceedings of the 33rd International Conference on International Conference on Machine Learning - Volume 48}, ser. ICML'16.\hskip 1em plus 0.5em minus 0.4em\relax JMLR.org, 2016, p. 40–48.

\bibitem{multi_prompt}
B.~Yao, G.~Chen, R.~Zou, Y.~Lu, J.~Li, S.~Zhang, Y.~Sang, S.~Liu, J.~Hendler, and D.~Wang, ``More samples or more prompts? exploring effective in-context sampling for llm few-shot prompt engineering,'' 2024.

\bibitem{prompt_recommend}
\BIBentryALTinterwordspacing
Z.~Yi, I.~Ounis, and C.~MacDonald, ``Contrastive graph prompt-tuning for cross-domain recommendation,'' \emph{ACM Trans. Inf. Syst.}, vol.~42, no.~2, dec 2023. [Online]. Available: \url{https://doi.org/10.1145/3618298}
\BIBentrySTDinterwordspacing

\bibitem{graphcl}
Y.~You, T.~Chen, Y.~Sui, T.~Chen, Z.~Wang, and Y.~Shen, ``Graph contrastive learning with augmentations,'' in \emph{Proceedings of the 34th International Conference on Neural Information Processing Systems}, ser. NIPS '20.\hskip 1em plus 0.5em minus 0.4em\relax Red Hook, NY, USA: Curran Associates Inc., 2020.

\bibitem{contrastive_backdoor}
\BIBentryALTinterwordspacing
H.~Zhang, J.~Chen, L.~Lin, J.~Jia, and D.~Wu, ``Graph contrastive backdoor attacks,'' in \emph{Proceedings of the 40th International Conference on Machine Learning}, ser. Proceedings of Machine Learning Research, A.~Krause, E.~Brunskill, K.~Cho, B.~Engelhardt, S.~Sabato, and J.~Scarlett, Eds., vol. 202.\hskip 1em plus 0.5em minus 0.4em\relax PMLR, 23--29 Jul 2023, pp. 40\,888--40\,910. [Online]. Available: \url{https://proceedings.mlr.press/v202/zhang23e.html}
\BIBentrySTDinterwordspacing

\bibitem{gnnguard}
X.~Zhang and M.~Zitnik, ``Gnnguard: Defending graph neural networks against adversarial attacks,'' in \emph{NeurIPS}, 2020.

\bibitem{random_backdoor}
\BIBentryALTinterwordspacing
Z.~Zhang, J.~Jia, B.~Wang, and N.~Z. Gong, ``Backdoor attacks to graph neural networks,'' in \emph{Proceedings of the 26th ACM Symposium on Access Control Models and Technologies}, ser. SACMAT '21.\hskip 1em plus 0.5em minus 0.4em\relax New York, NY, USA: Association for Computing Machinery, 2021, p. 15–26. [Online]. Available: \url{https://doi.org/10.1145/3450569.3463560}
\BIBentrySTDinterwordspacing

\bibitem{spectral_backdoor}
X.~Zhao, H.~Wu, and X.~Zhang, ``Effective backdoor attack on graph neural networks in spectral domain,'' \emph{IEEE Internet of Things Journal}, vol.~11, no.~7, pp. 12\,102--12\,114, 2024.

\bibitem{motif_backdoor}
H.~Zheng, H.~Xiong, J.~Chen, H.~Ma, and G.~Huang, ``Motif-backdoor: Rethinking the backdoor attack on graph neural networks via motifs,'' \emph{IEEE Transactions on Computational Social Systems}, vol.~11, no.~2, pp. 2479--2493, 2024.

\end{thebibliography}

\appendix

\section*{Empirical Studies}
\label{appendix_dataset}

\subsection{Dataset Statistics}
We leverage $4$ real-world graph datasets that are extensively evaluated for node classification tasks, including 1) Cora and CiteSeer \cite{cora_citeseer}, where each node denotes a document with a bag-of-words representation, and each edge represents a citation relationship; 2) Physics \cite{physics_computers}, where each node represents an author, the node feature contains paper keywords for each author's papers and each edge indicates authors' collaborations; 3) Computers \cite{physics_computers}, where each node represents a product, each node feature is product review encoded by a bag-of-words model, and each edge indicates that two products are frequently purchased together. The detailed graph statistics are shown in Table \ref{data_stat}.

\vspace{-5mm}
\begin{table}[!htbp]
\caption{\centering Detailed statistics of four graph datasets.}
\resizebox{\linewidth}{!}{
\begin{tabular}{c|cccc|cc}
\toprule
\textbf{Datasets}  & \textbf{Nodes}  & \textbf{Edges}   & \textbf{Features} & \textbf{Labels} & \textbf{Avg. Deg} & \textbf{Homo.(\%)} \\ \hline
\textbf{Cora}      & 2,708  & 10,556  & 1,433    & 7      & 2.65            & 81.40          \\
\textbf{CiteSeer}  & 3,327  & 9,104   & 3,703    & 6      & 2.74            & 70.62          \\
\textbf{Physics}   & 34,493 & 495,924 & 8,415    & 5      & 14.38            & 91.53          \\
\textbf{Computers} & 13,752 & 491,722 & 767      & 10     & 35.76            & 78.53          \\ \bottomrule
\end{tabular}}
\label{data_stat}
\end{table}

\vspace{-5mm}
\subsection{Attack Pipeline}
\label{attack_pipe}
The attack pipeline is summarized as follows: first, to ensure almost $100\%$ ASRs and assess whether graph prompts can reveal more vulnerabilities under such extreme conditions, we assume that attackers randomly poison $50\%$ of training nodes with the victim label. The overall PRs range from $10.53\%$ to $25.26\%$ over the entire training samples in different graphs. Next, attackers flip the labels of poisoned candidates to the target label. Attackers then either employ All-in-One to generate universal triggers for these poisoned candidates, or inject the hand-crafted triggers mentioned in Sun et al. \cite{all_in_one}. Finally, during the inference phase, these triggers are attached to half of the test nodes. Please refer to the attack performance of eight transfer cases in Table \ref{pre_exp}.

\vspace{-2mm}
\subsection{Benign and attack performance of graph prompt tuning}
We select the small-scale datasets, Cora and Citeseer, as the target domains, and all four graphs as the source domains. 

\vspace{-5mm}
\begin{table}[h]
\caption{Performance of benign and trigger-injected graph prompt tuning, We bold the best performance.}
\resizebox{\linewidth}{!}{
\begin{tabular}{c|cccc|cccc}
\toprule
\textbf{Source\textbackslash{}Target} & \multicolumn{4}{c|}{\textbf{Cora}} & \multicolumn{4}{c}{\textbf{CiteSeer}} \\ \hline
(Metrics)                                     & ACC     & F1      & AUC     & ASR     & ACC     &F1     & AUC       & ASR    \\ \hline
\textbf{Cora}                                 & \textbf{81.03}         & \textbf{69.55}        & \textbf{95.64}      & \textbf{100.00}      &   78.74       & 68.38     & \textbf{91.08}    & 99.72 \\
\textbf{CiteSeer}                             & 71.25        & 53.24        & 89.52      & \textbf{100.00}      & \textbf{81.86}         & \textbf{80.83}     & 87.85    & \textbf{100.00} \\ 
\textbf{Physics}                              & 69.04        & 63.65        & 88.14      & \textbf{100.00}      & 79.94         & 78.76     & 90.35    & 99.72 \\ 
\textbf{Computers}                            & 30.15        & 6.62        & 51.88      & \textbf{100.00}      & 70.03    & 62.58    & 88.51    & \textbf{100.00}   \\ 
\bottomrule
\end{tabular}}
\vspace{-3mm}
\label{pre_exp}
\end{table}

\subsection{Node-centric Homophily}
\label{homo_appendix}
The local-subgraph and global-view homophily distributions of the whole training graph in the CiteSeer to Cora transfer are shown in Figure \ref{graph_homo_citeseer_cora}. Since there exist eight transfer cases, we only present the homophily distributions of the poisoned samples with Cora as the target domain in Figures \ref{graph_homo_cora_cora}-\ref{graph_homo_computers_cora}. As depicted in Section \ref{sec:prelim_exp}, except for the negative transfer from Computers to Cora, the rest of transfers exhibit the consistent patterns.
\begin{figure}[H]
    \centering
    \includegraphics[width=\linewidth]{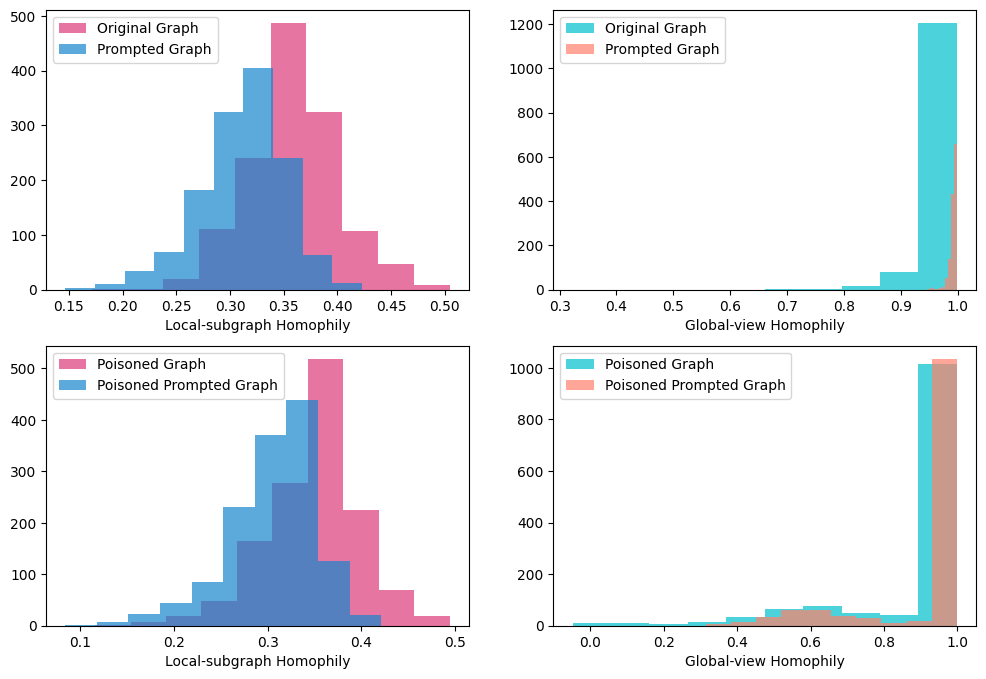}
    \caption{Local-subgraph(LHS) and global-view(RHS) homo- phily distributions of the whole training graph in the CiteS- eer to Cora transfer.}
    \label{graph_homo_citeseer_cora}
    \vspace{-6mm}
\end{figure}

\begin{figure}[ht]
    \centering
    \includegraphics[width=\linewidth]{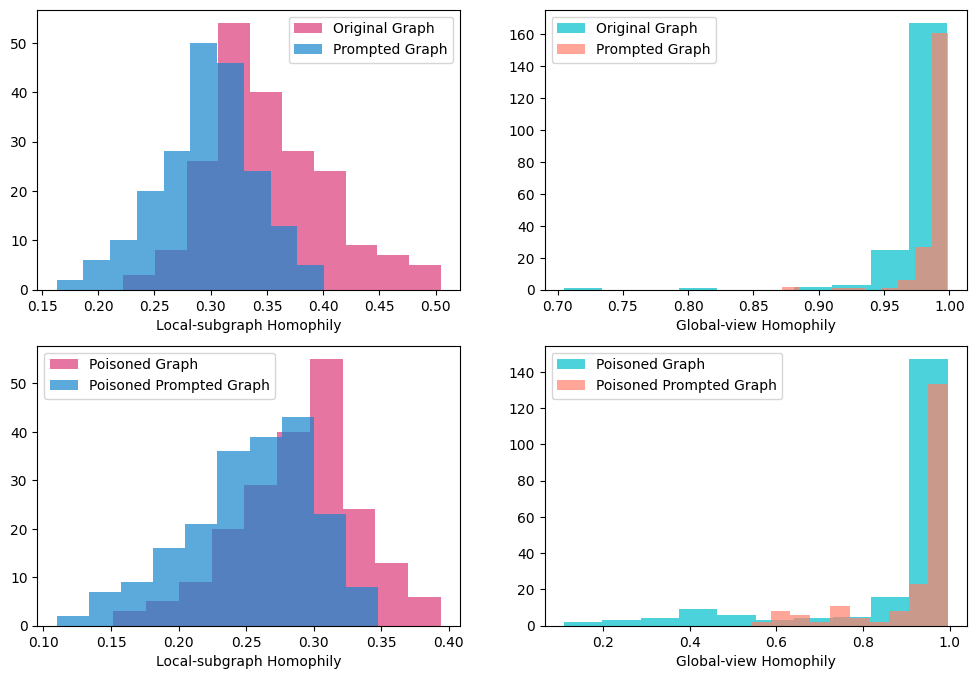}
    \caption{Local-subgraph(LHS) and global-view(RHS) homo- phily distributions of poisoned nodes in the Cora to Cora tr- ansfer.}
    \label{graph_homo_cora_cora}
    \vspace{-6mm}
\end{figure}

\begin{figure}[ht]
    \centering
    \includegraphics[width=\linewidth]{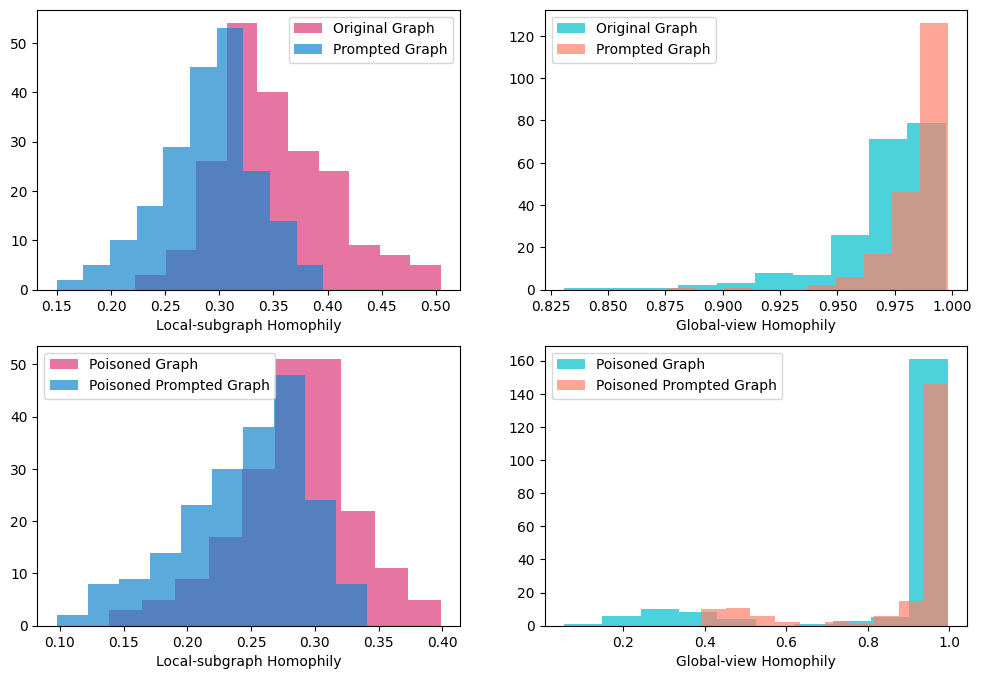}
    \caption{Local-subgraph(LHS) and global-view(RHS) homo- phily distributions of poisoned nodes in the Physics to Cora transfer.}
    \label{graph_homo_physics_cora}
    \vspace{-6mm}
\end{figure}

\begin{figure}[ht]
    \centering
    \includegraphics[width=\linewidth]{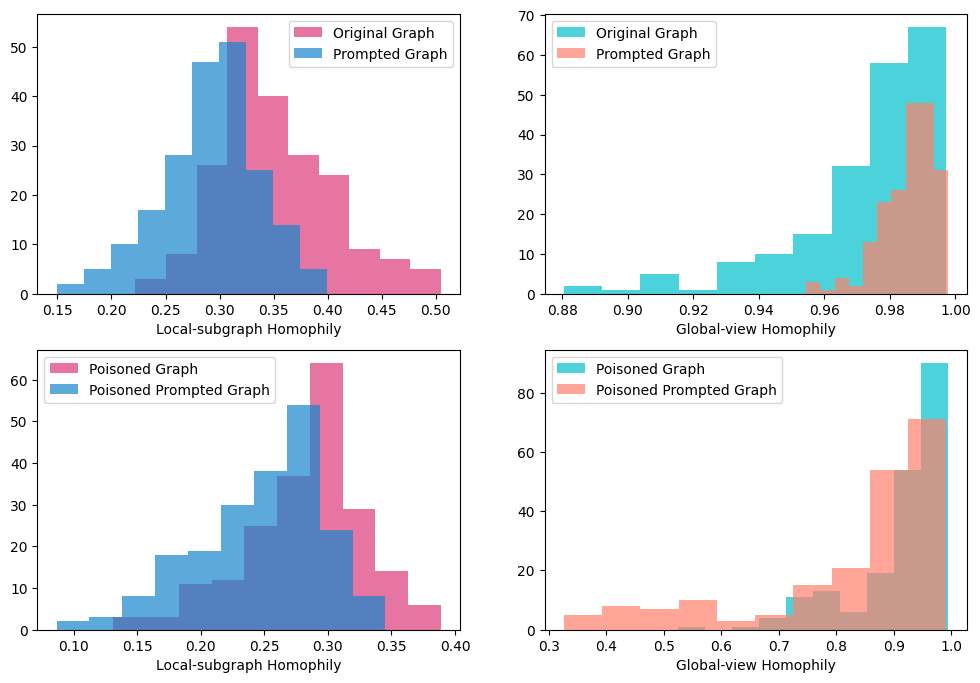}
    \caption{Local-subgraph(LHS) and global-view(RHS) homo- phily distributions of poisoned nodes in the Computers to C- ora transfer.}
    \label{graph_homo_computers_cora}
\end{figure}


\subsection{Centroid Similarity}
\label{centroid_sim_appendix}
Since there exist eight transfer cases, we only present the centroid alignment, misalignment and difference distributions of the poisoned samples with Cora as the target domain in Figures \ref{graph_sim_mis_cora_cora}-\ref{graph_sim_mis_computers_cora}. As depicted in Section \ref{sec:prelim_exp}, except for the negative transfer from Computers to Cora, the rest of transfers render the consistent patterns.

\begin{figure}[H]
    \centering
    \includegraphics[width=\linewidth]{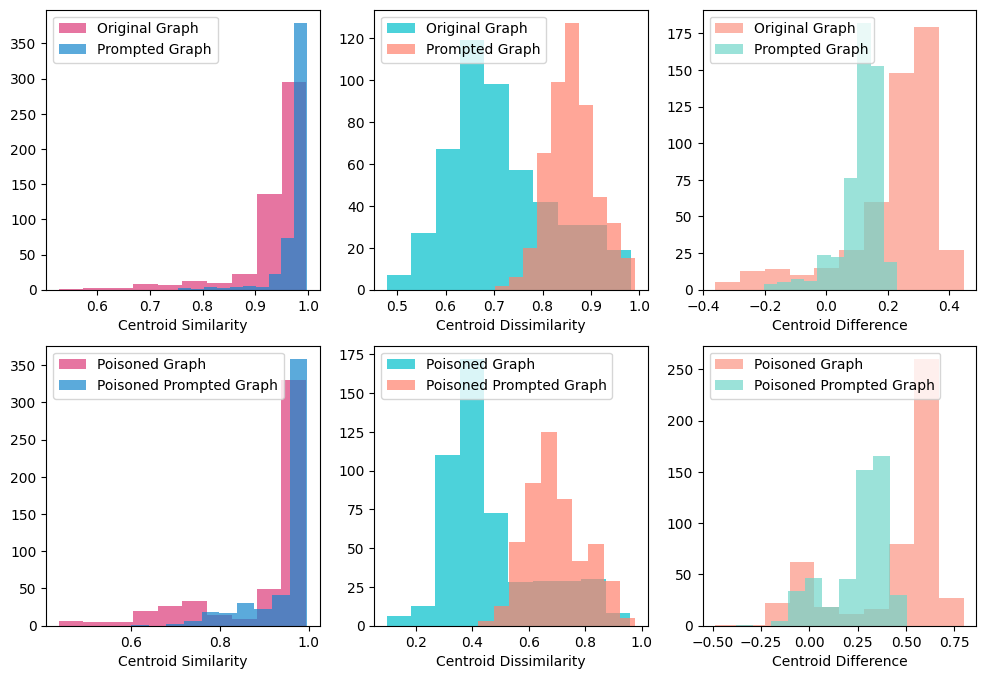}
    \caption{Centroid alignment(LHS), misalignment(Center) and difference(RHS) distributions of poisoned nodes in the Cora to Cora transfer.}
    \label{graph_sim_mis_cora_cora}
\end{figure}

\begin{figure}[ht]
    \centering
    \includegraphics[width=\linewidth]{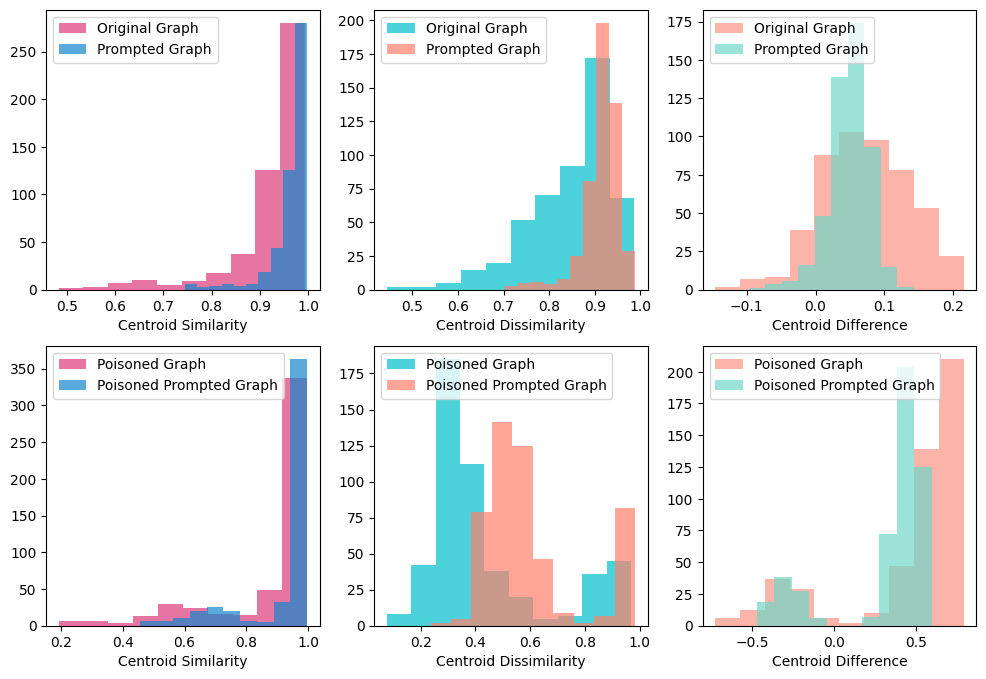}
    \caption{Centroid alignment(LHS), misalignment(Center) and difference(RHS) distributions of poisoned nodes in the Physi- cs to Cora transfer.}
    \label{graph_sim_mis_physics_cora}
    \vspace{-6mm}
\end{figure}

\begin{figure}[ht]
    \centering
    \includegraphics[width=\linewidth]{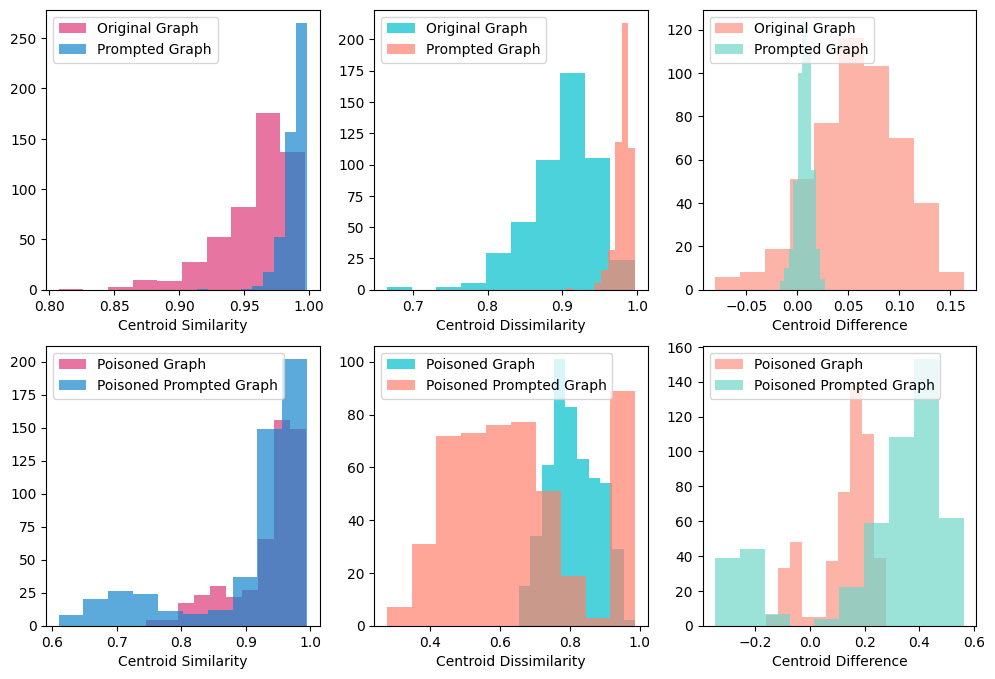}
    \caption{Centroid alignment(LHS), misalignment(Center) and difference(RHS) distributions of poisoned nodes in the Com- puters to Cora transfer.}
    \label{graph_sim_mis_computers_cora}
    \vspace{-6mm}
\end{figure}

\vspace{-6mm}
\subsection{Embedding Space Visualization}
\label{emb_space_appendix}

Similarly, we present the node embedding visualization with Cora as the target domain in Figures \ref{emb_space}-\ref{graph_emb_computers_cora}. As depicted in Section \ref{sec:prelim_exp}, except for the negative transfer from Computers to Cora, the rest of transfers render the consistent patterns.

\begin{figure}[H]
    \centering
    \includegraphics[width=\linewidth]{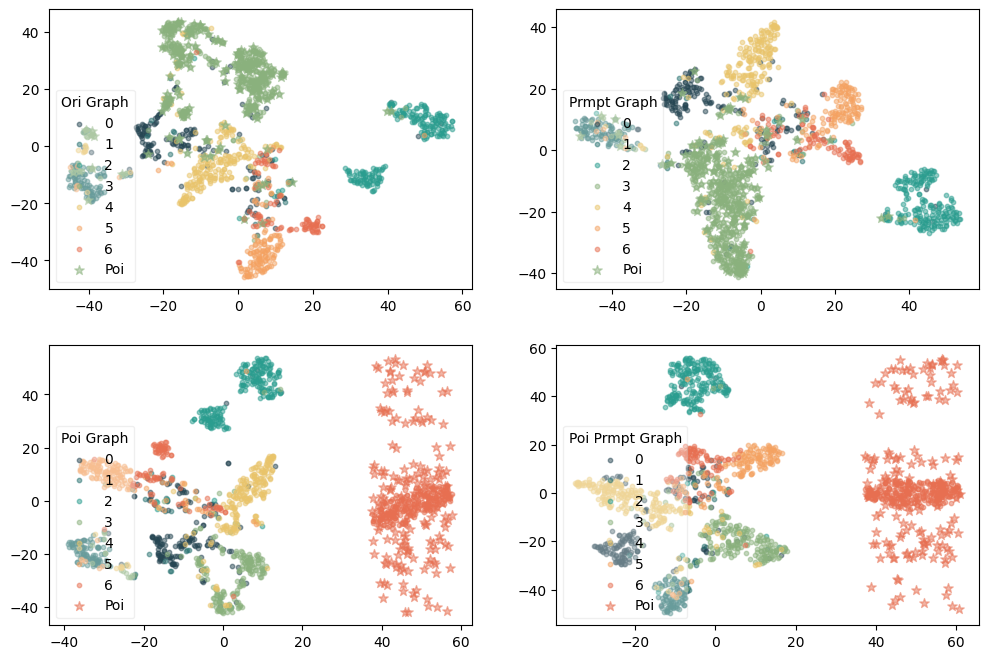}
    \caption{Node embeddings. The top two subfigures denote node embeddings of the original graph and prompted graph, where the lawngreen stars represent poisoned nodes; the bottom two denote the poisoned graph and poisoned prompted graph, where the red stars represent poisoned nodes.}
    \label{emb_space}
    \vspace{-6mm}
\end{figure}

\begin{figure}[ht]
    \centering
    \includegraphics[width=\linewidth]{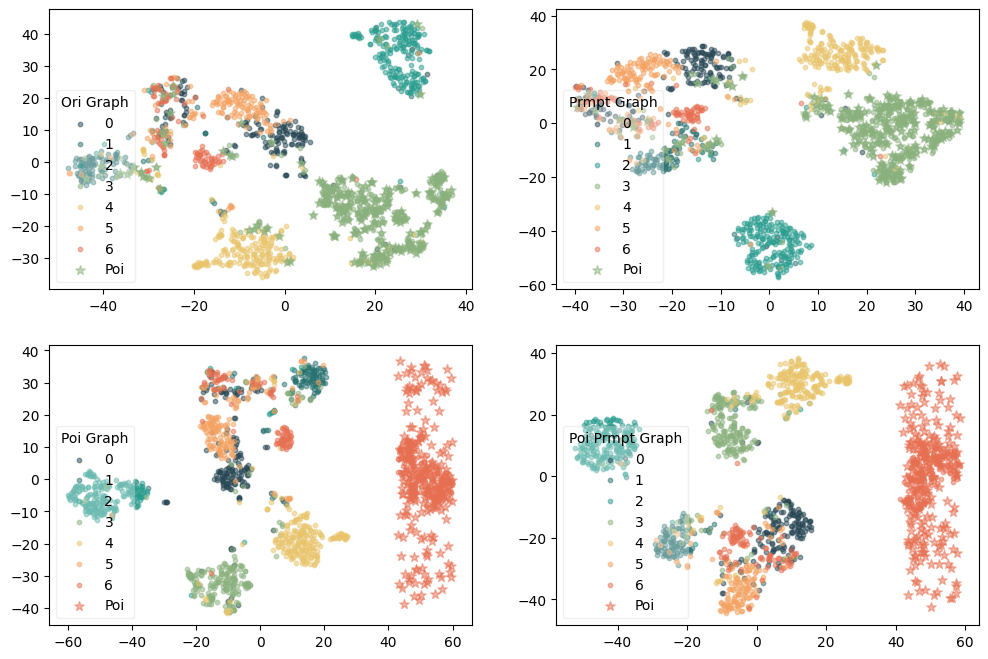}
    \caption{\centering{Node embeddings in the Cora to Cora transfer.}}
    \label{graph_emb_cora_cora}
    \vspace{-6mm}
\end{figure}

\begin{figure}[ht]
    \centering
    \includegraphics[width=\linewidth]{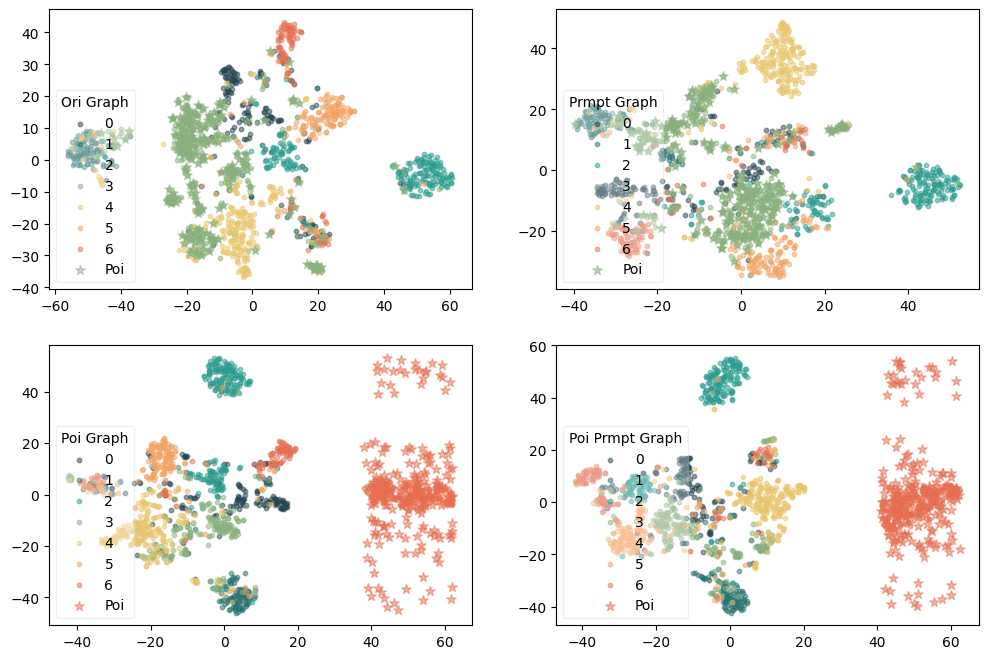}
    \caption{\centering{Node embeddings in the Physics to Cora transfer.}}
    \label{graph_emb_physics_cora}
    \vspace{-6mm}
\end{figure}

\begin{figure}[ht]
    \centering
    \includegraphics[width=\linewidth]{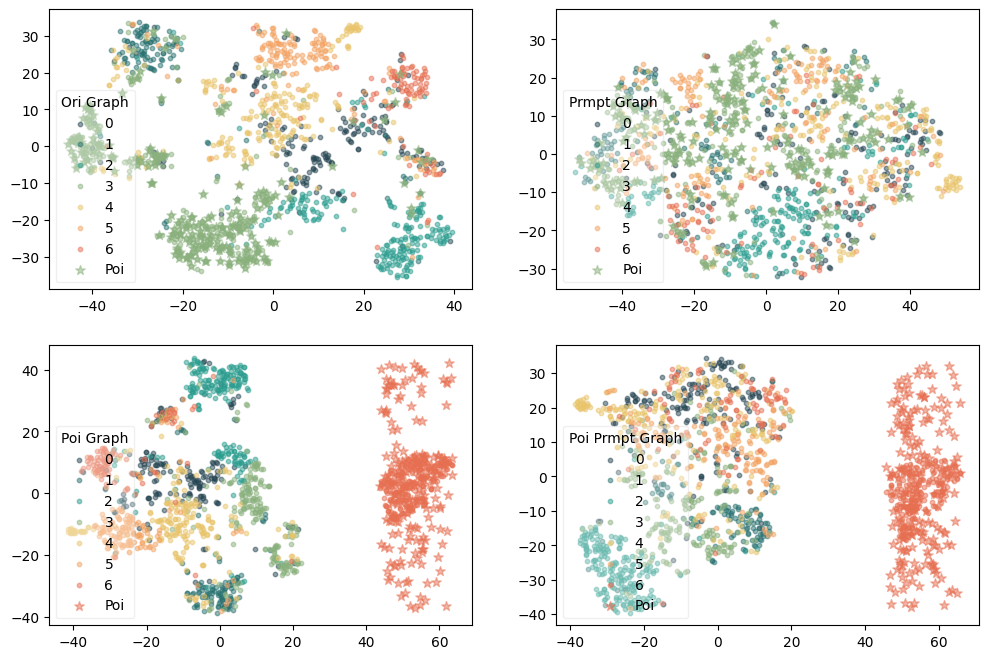}
    \caption{\centering{Node embeddings in the Computers to Cora transfer.}}
    \label{graph_emb_computers_cora}
    \vspace{-6mm}
\end{figure}

\section*{Attack Framework}
\subsection{Algorithms}
\label{sec:algo}

In this section, we present the overview algorithm of poisoned candidate selection based on different attack types and how to train Krait under diverse trigger generation methods.

\subsubsection{Algorithm 1 - Poisoned Candidate Selection}
\label{algo1}
\begin{breakablealgorithm}
\caption{Poisoned Candidate Selection}\label{alg_poison}
\begin{algorithmic}[1]
\Require Training graph $\mathcal{G}_{train}$, poisoning rate $p$, pre-defined degree threshold $d_{pre}$, ground-truth label $Y$, current label $Y_{c}$, target label $Y_{t}$, attack type$m$, index searching function idx($\cdot$), choosing top k function topk($\cdot$), ranking function rnk($\cdot$)
\Ensure Poisoned candidates $\mathcal{B}$
\Function{One2one}{$\mathcal{G}_{train}, p, d_{pre}, Y_c$}
    \For{each $i \in \mathcal{G}_{train}$}
        \If{$Y_i=Y_c \ \& \ d(i)\leq d_{pre}$}
            \State $\mathcal{G}_{cur} \leftarrow \mathcal{G}_{train}$
        \EndIf
    \EndFor
    \State Compute $LNH_{cur}$ as in Equation \ref{LNH}
    \State $r_{cur}=rnk(LNH_{cur})$
    \State $b=topk(r_{cur}), k=p\times|\mathcal{V}_{cur}|$ 
    \State $\mathcal{B}=idx(b)$
    
    \Return $\mathcal{B}$
\EndFunction
\If{$m=$one-to-one}
    \State Implement $ONE2ONE$, return $\mathcal{B}$
    \State $Y_{\mathcal{B}} \leftarrow Y_t$
    \State $\mathcal{G}\prime_{train} \leftarrow \mathcal{G}_{train}$
\ElsIf{$m=$all-to-one}
    \State $\mathcal{B} \leftarrow \emptyset$
    \For{$j \in |Y|$}
        \If{$j\neq Y_t$}
            \State $\mathcal{B}_{j}=ONE2ONE(\mathcal{G}_{train}, p, d_{pre}, j)$
            \State $\mathcal{B}=\mathcal{B}\cup\mathcal{B}_{j}$
        \EndIf
    \EndFor
    \State $Y_{\mathcal{B}} \leftarrow Y_t$
    \State $\mathcal{G}\prime_{train} \leftarrow \mathcal{G}_{train}$
\ElsIf{$m=$all-to-all}
    \State $\mathcal{B} \leftarrow \emptyset$
    \For{$j \in |Y|$}
        \State $\mathcal{B}_{j}=ONE2ONE(\mathcal{G}_{train}, p, d_{pre}, j)$
        \State $\mathcal{B}=\mathcal{B}\cup\mathcal{B}_{j}$
        \State $Y\prime_t \leftarrow (j+1)mod|Y|$
        \State $Y_{\mathcal{B}_{j}} \leftarrow Y\prime_t$
    \EndFor
    \State $\mathcal{G}\prime_{train} \leftarrow \mathcal{G}_{train}$
\Else
    \State break
\EndIf \\
\Return $\mathcal{B}, \ \mathcal{G}\prime_{train}$
\end{algorithmic}
\end{breakablealgorithm}

\subsubsection{Algorithm 2 - Krait} 
\label{algo2}
\begin{breakablealgorithm}
\caption{Krait}\label{alg_krait}
\begin{algorithmic}[1]
\Require Training graph $\mathcal{G}_{train}$, poisoning rate $p$, pre-defined degree threshold $d_{pre}$, ground-truth label $Y$, current label $Y_{c}$, target label $Y_{t}$, attack type$m$, trigger generation method $\rho$, poisoned warm-up candidates $\mathcal{G}_{\mathcal{P}}$, default parameters dft, training process opt$(\cdot)$
\Ensure Backdoored graph prompt tuning model $f\prime_{prm}$
\State Implement Algorithm \ref{alg_poison}
\If{$\rho=$invoke}
    \State Initialize $g_{\psi}$
    \State Implement $f_{prm}$(dft, $\mathcal{G}_{\mathcal{P}}$), return poisoned $g\prime_{\psi}$
    \State $\mathcal{G}\prime_{train} \leftarrow g\prime_{\psi}(\mathcal{G}\prime_{train})$
    \For{$i \in (1, epochs)$}
        \State $f\prime_{prm}=opt(dft, f_{prm}, g_{\psi}, \mathcal{G}\prime_{train}, \mathcal{L}_{bkd})$
    \EndFor
\ElsIf{$m=$interact}
    \State Initialize $g_{\psi}$, $g\prime_{\psi}$
    \For{$i \in (1, epochs)$}
        \State $\mathcal{G}\prime_{train} \leftarrow g\prime_{\psi}(\mathcal{G}\prime_{train})$
        \State $\mathcal{G}\prime_{train} \leftarrow g_{\psi}(\mathcal{G}\prime_{train})$
        \State $f\prime_{prm}=opt(dft, f_{prm}, g_{\psi}, g\prime_{\psi}, \mathcal{G}\prime_{train}, \mathcal{L}_{bkd})$
    \EndFor
\ElsIf{$m=$modify}
    \State Initialize $g_{\psi}$, $g\prime_{\psi}$
    \For{$i \in (1, epochs)$}
        \State $\mathcal{G}\prime_{train} \leftarrow g_{\psi}(\mathcal{G}\prime_{train})$
        \State $\mathcal{G}\prime_{train} \leftarrow g\prime_{\psi}(\mathcal{G}\prime_{train})$
        \State $f\prime_{prm}=opt(dft, f_{prm}, g_{\psi}, g\prime_{\psi}, \mathcal{G}\prime_{train}, \mathcal{L}_{bkd})$
    \EndFor
\Else
    \State break
\EndIf
\Return $f\prime_{prm}$
\end{algorithmic}
\end{breakablealgorithm}

\vspace{1mm}
\section*{Experiments}
\subsection{Implementation Details}
\label{imp_details}
We follow the default settings of All-in-One \cite{all_in_one}. We set the number of graph neural layers as $2$ with a hidden dimension of $100$ and leverage SVD (Singular Value Decomposition) to reduce the initial features to $100$ dimensions. For pre-training, the learning rate is set to $0.01$, weight decay is $1e-4$, and epoch is $100$. For prompt tuning, the cross-pruning rate is set to $0.1$ and the inner-pruning rate is $0.3$. We utilize cross-entropy loss throughout our experiments. And batch size is $10$. For more inner parameters, please refer to All-in-One's GitHub repository: \url{https://github.com/sheldonresearch/ProG/tree/main}. 
\subsection{Benign Performance of Graph Prompt Tuning}
\label{benign_perf_app}
\begin{table*}[htbp]
\caption{Performance of benign graph prompt tuning. The row/column denotes the source/target domains, respectively. We bold the best performance in each transfer case.}
\resizebox{\linewidth}{!}{
\begin{tabular}{c|ccc|ccc|ccc|ccc}
\toprule
\textbf{Src\textbackslash{}Tar}  & \multicolumn{3}{c|}{\textbf{Cora}} & \multicolumn{3}{c|}{\textbf{CiteSeer}} & \multicolumn{3}{c|}{\textbf{Physics}} & \multicolumn{3}{c}{\textbf{Computers}} \\ \hline
(Metrics) & ACC     & F1      & AUC   & ACC      & F1      & AUC      & ACC      & F1      & AUC     & ACC      & F1      & AUC      \\ \hline
\textbf{Cora}      & 82.22   & 73.97  & 95.98 & 80.76        & 73.99        & \textbf{91.77}         & \textbf{87.58}         & \textbf{82.51}        & \textbf{94.20}        & 56.46         & 25.62        & 66.18         \\
\textbf{CiteSeer}  & \textbf{86.85}   & \textbf{84.96}   & \textbf{96.63} & \textbf{83.07}        & \textbf{82.37}        & 89.12         & 82.70         & 64.67        & 85.91       & 50.02         & 17.48        & 62.22         \\
\textbf{Physics}   & 72.04  & 59.68 & 88.85 & 80.32        & 77.46        & 91.54         & 79.94         & 67.11        & 87.59        & \textbf{61.92}         & \textbf{28.34}        & \textbf{72.66}         \\
\textbf{Computers} & 47.10  & 25.25 & 67.63   & 54.50       & 40.01        & 78.63         & 79.73         & 59.26       & 81.14       &47.44         & 17.64        & 60.79         \\ \bottomrule
\end{tabular}}
\label{benign_perf}
\end{table*}

\subsection{Effectiveness: Bad-case Analysis}
\label{bad_effective}

For all-to-all attacks, we choose the victim and target label attack pair based on $Y_t = (y + 1)mod|\mathcal{Y}|$ \cite{all_to_one_all}, where the target label will be automatically chosen as the next label after the victim label in the label sequence. Therefore we would have $|\mathcal{Y}|$ attack pairs in total. However, some attack pairs’ approximated decision boundaries might not be captured by label non-uniformity homophily (LNH). For instance, if a node’s label is class 0 (its target label is class 1 due to the strategy), its neighbors’ labels are almost class 5, hence, this node might have a higher LNH value and can be marked as one of the trigger positions by Krait. This node is easier to be misclassified as class 5 instead of its target label - class 1 in the attack pair. Due to this strategy, Krait might not outperform the baselines in some cases, but still excels in large-scale transfer cases, i.e., Physics to CiteSeer transfer and Physics to Computers transfer. We believe that one of the promising directions of future work lie in adaptively selecting the attack pairs based on LNH and LDN. For instance, in the aforementioned example, we can choose class 5 as this node's target label. Moreover, all-to-all attacks force graph prompt backdoors to simultaneously learn multiple mapping functions among triggers and target labels. However, as graph prompt tuning only adaptively tunes pre-trained models within a few steps, it is highly challenging to learn multiple solid mapping functions within insufficient training, leading to relatively lower performance in all-to-all attacks.

\subsection{Flexibility of Krait}
\label{flex_perf}

\begin{table*}[ht]
\caption{\centering{Performance of Krait under different GNN backbones. Cite2Cora: CiteSeer to Cora transfer.}}
\resizebox{\linewidth}{!}{
\begin{tabular}{c|c|c|cccc|cccc|cccc}
\toprule
\textbf{Transfer}                  & \multirow{3}{*}{\textbf{Backbone}} & \multirow{3}{*}{\textbf{Model}} & \multicolumn{4}{c|}{\textbf{One-to-One(Invoke)}}                                                      & \multicolumn{4}{c|}{\textbf{All-to-One(Interact)}}                                                      & \multicolumn{4}{c}{\textbf{All-to-All(Modify)}}                                                      \\ \cline{1-1} \cline{4-15} 
    (Evaluation)                   &                           &                        & \multicolumn{2}{c|}{\textbf{Effectiveness}} & \multicolumn{2}{c|}{\textbf{Stealthiness}} & \multicolumn{2}{c|}{\textbf{Effectiveness}} & \multicolumn{2}{c|}{\textbf{Stealthiness}} & \multicolumn{2}{c|}{\textbf{Effectiveness}} & \multicolumn{2}{c}{\textbf{Stealthiness}} \\ \cline{1-1} \cline{4-15} 
(Metrics)                      &                           &                        & ASR       & \multicolumn{1}{c|}{AMC}      & CA                    & AHD        & ASR       & \multicolumn{1}{c|}{AMC}      & CA                   & AHD         & ASR       & \multicolumn{1}{c|}{AMC}      & CA                  & AHD         \\ \hline
\multirow{9}{*}{\textbf{Cite2Cora}} & \multirow{3}{*}{\textbf{GCN}}   & \textbf{Random}                 & 16.00          & 82.91                              & 87.70                         & 6.79            & 98.24          & \textbf{99.12}                              & 86.57                        & 5.25            & \textbf{31.25}          & 93.77                              & \textbf{86.75}                        & 4.58            \\
                               &                           & \textbf{LNH}                    & 30.67          & 84.36                              & \textbf{87.73}                         & \textbf{6.62}            & \textbf{98.59}          & 99.10                              & 86.77                         & 5.72            & 27.75          & 90.12                              & 86.54                        & \textbf{3.94}            \\
                               &                           & \textbf{Full}                  & \textbf{86.89}          & \textbf{95.75}                              & \textbf{87.73}                         & 9.94            & 94.13          & 98.70                              & 86.45                         & 6.63            & 26.81          & \textbf{94.49}                              & 86.36                       & 4.13            \\ \cline{2-3}
                               & \multirow{3}{*}{\textbf{GAT}} & \textbf{Random}                 & 60.44          & 93.32                              & 69.52                         & \textbf{0.59}            & 99.38          & 99.97                              & 66.33                        &  \textbf{4.94}           & \textbf{16.42}          & 91.15                              & \textbf{68.13}                      & 5.06            \\
                               &                           & \textbf{LNH}                    & 66.67          & 94.73                              & 71.67                      & 0.63            & 99.21         & 99.95                              & 67.30                     & 5.16            & 16.24          & \textbf{93.32}                              & 67.72                   & 5.52            \\
                               &                           & \textbf{Full}                  & \textbf{100.00}          & \textbf{99.80}                              & \textbf{71.91}                       & 3.34            & \textbf{100.00}          & \textbf{100.00}                              & \textbf{68.57}                     & 5.44            & 13.63          & 89.35                              & 66.22                   & \textbf{4.96}            \\ \cline{2-3}
                               & \multirow{3}{*}{\textbf{GT}}   & \textbf{Random}                 &94.22               & 93.12                         & 86.89                            & \textbf{0.18}             & 99.91 & 99.75            & 85.66   & \textbf{3.65}             & \textbf{42.61}               & \textbf{93.30}                         & 85.83                             & 6.38            \\
                               &                           & \textbf{LNH}                    & 95.56               & 93.20                         & \textbf{86.95}                            & 0.19             & 95.51  & 98.86           & 85.68  & 3.71             & 38.68               & 89.56                         & 85.83                             & 6.13             \\
                               &                           & \textbf{Full}                  & \textbf{100.00}               & \textbf{100.00}                         & 86.75                             & 3.02             & \textbf{100.00}               & \textbf{100.00}                         & \textbf{85.86}                            & 4.05             & 31.89               & 89.05                         & \textbf{86.27}                            & \textbf{5.80}             \\ \bottomrule
\end{tabular}}
\label{backbone_Krait}
\end{table*}

\subsection{Black-box Setting: Bad-case Analysis}
\label{bad_black}
For the black-box setting in Table \ref{black_Krait}, when the backbone of the victim model closely resembles that of the open-sourced model (albeit with different parameters), the performance largely improves. We speculate that the lower performances of some cases stem from mismatching issues between the victim model and learnable triggers that are generated from different GNN architectures. All-to-one attacks under the black-box setting stay ahead, as they in essence enlarge the poisoned samples after label flipping to learn more solid mappings between triggers and target labels than one-to-one attacks. However, all-to-all attacks still struggle to learn multiple solid mappings simultaneously within a few training steps.

\subsection{Backdoored Node Embedding Visualization}
\label{Backdoored_emb_viz_appendix}
\begin{figure}[H]
    \centering
    \subfloat[One-to-one attack]{  \includegraphics[width=0.4832\linewidth]{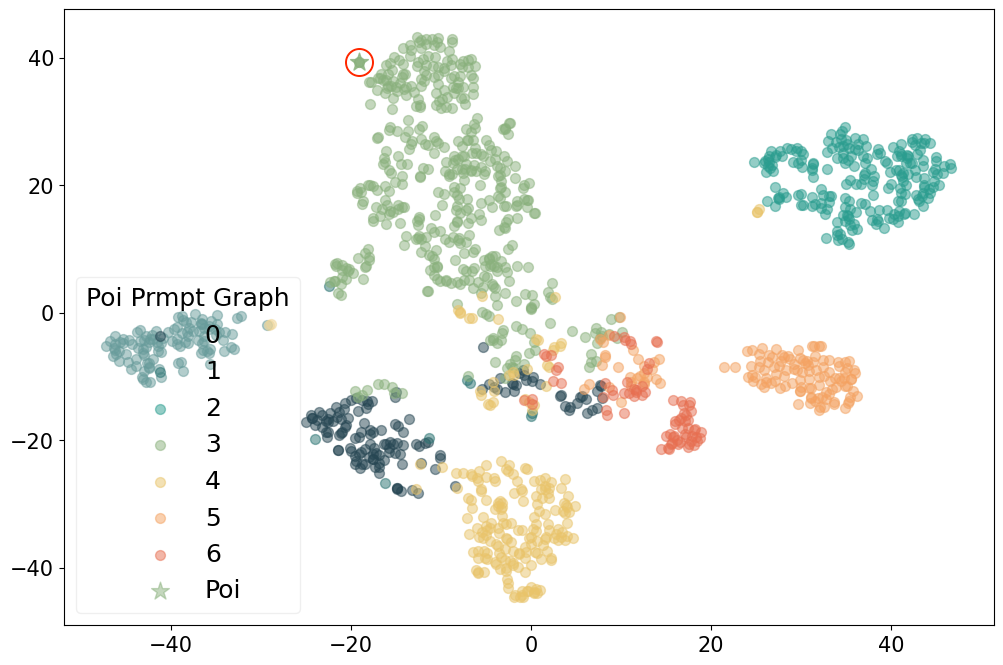}
    \label{one-to-one}}
    \subfloat[All-to-one attack]{   \includegraphics[width=0.4832\linewidth]{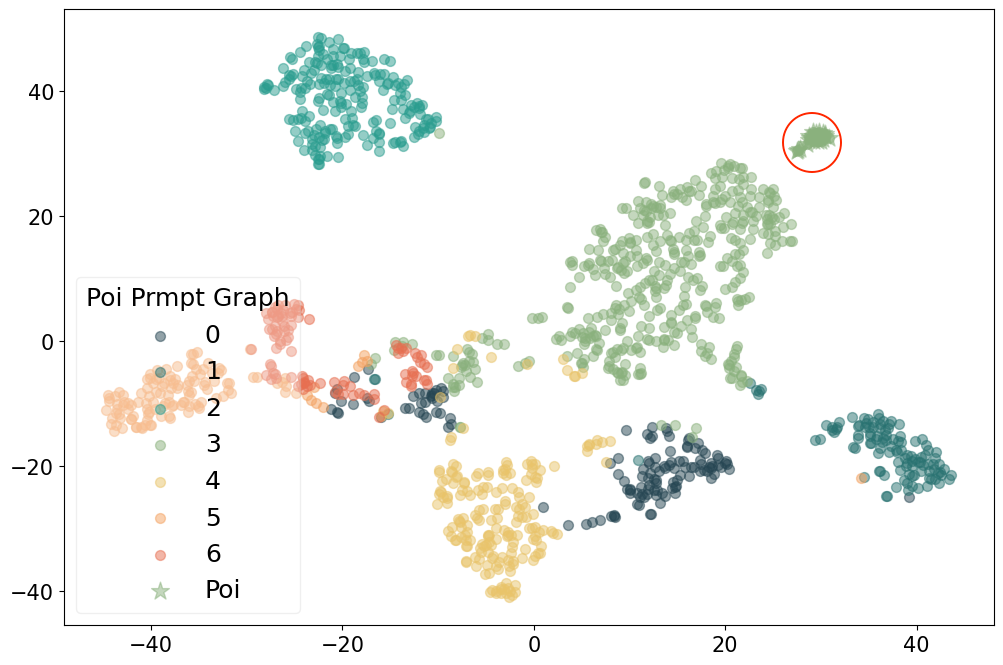}
    \label{all-to-one}}
    \caption{Backdoored node embedding of ``Invoke", poisoned nodes are marked with red circles.}
    \label{prompt_backdoor_viz}
\end{figure}


\begin{figure}[H]
    \centering
    \subfloat[\centering One-to-one attack]{  \includegraphics[width=0.4832\linewidth]{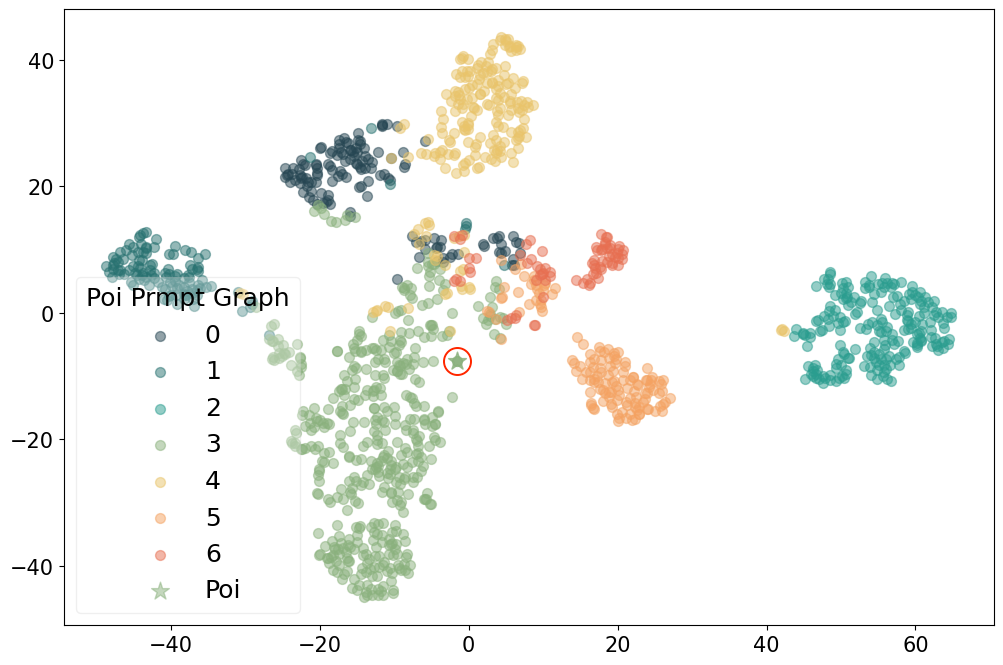}
    \label{interact_one-to-one_appendix}}
    \subfloat[\centering All-to-one attack]{   \includegraphics[width=0.4832\linewidth]{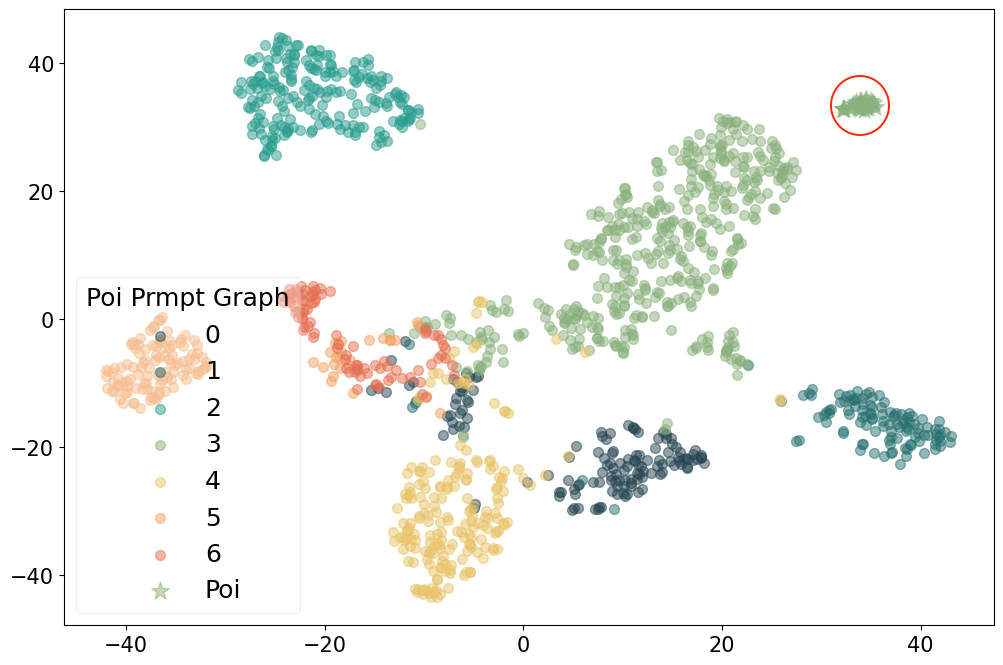}
    \label{interact_all-to-one_appendix}}
    \caption{Prompted node embedding with the ``Interact" inject- ion.}
    \label{prompt_backdoor_viz_interact_appendix}
\end{figure}

\begin{figure}[htbp]
    \centering
    \subfloat[\centering One-to-one attack]{  \includegraphics[width=0.4832\linewidth]{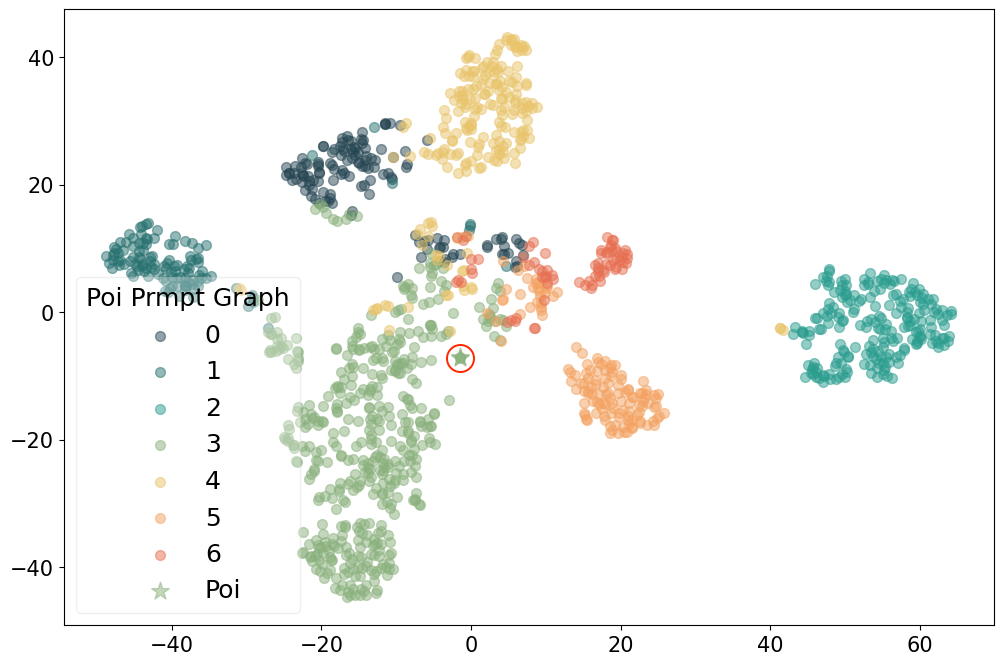}
    \label{modify_one-to-one_appendix}}
    \subfloat[\centering All-to-one attack]{   \includegraphics[width=0.4832\linewidth]{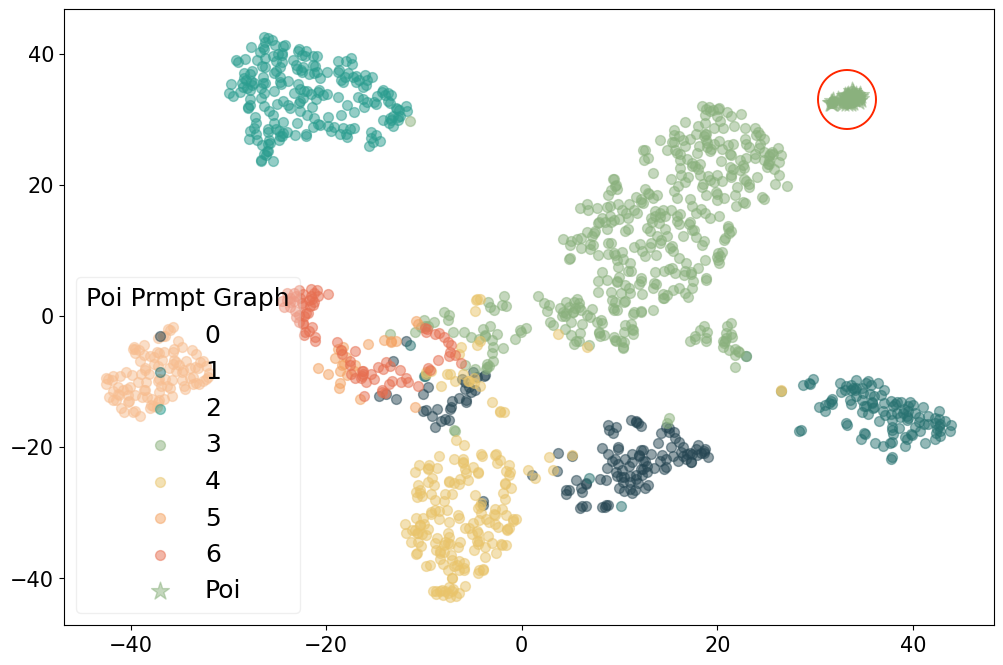}
    \label{modify_all-to-one_appendix}}
    \caption{Prompted node embedding with the ``Modify" inject- ion.}
    \label{prompt_backdoor_viz_modify_appendix}
\end{figure}

As shown in Figure \ref{prompt_backdoor_viz}-\ref{prompt_backdoor_viz_modify_appendix}, the projections of poisoned samples in different trigger generation methods exhibit similar patterns, where these samples are at the edge of class boundaries, regardless of attack types. 

\end{document}